\documentclass[runningheads]{llncs}

\usepackage{eccv}

\usepackage{eccvabbrv}

\usepackage{graphicx}
\usepackage{booktabs}

\usepackage[accsupp]{axessibility}  

\usepackage{color}
\usepackage{epsfig}
\usepackage{graphicx}

\usepackage{adjustbox}
\usepackage{array}
\usepackage{booktabs}
\usepackage{colortbl}
\usepackage{wrapfig}
\usepackage{hhline}
\usepackage{multirow}
\usepackage{wrapfig}
\usepackage{floatflt}
\usepackage{mwe}
\usepackage{float}
\usepackage{placeins}

\usepackage{amsmath,amsfonts,amssymb}
\usepackage{mathtools}  
\usepackage{bm}
\usepackage{nicefrac}
\usepackage{microtype}

\usepackage{changepage}
\usepackage{extramarks}
\usepackage{fancyhdr}
\usepackage{lastpage}
\usepackage{setspace}
\usepackage{soul}
\usepackage{xspace}

\usepackage{enumerate}
\usepackage{enumitem}  

\usepackage{makecell}

\usepackage{pifont} 

\usepackage{algorithm,algpseudocode}

\usepackage[symbol]{footmisc}

\usepackage{afterpage}

\newcolumntype{L}[1]{>{\raggedright\let\newline\\\arraybackslash\hspace{0pt}}m{#1}}
\newcolumntype{C}[1]{>{\centering\let\newline\\\arraybackslash\hspace{0pt}}m{#1}}
\newcolumntype{R}[1]{>{\raggedleft\let\newline\\\arraybackslash\hspace{0pt}}m{#1}}

\newcommand{\ignore}[1]{}

\makeatletter
\DeclareRobustCommand\onedot{\futurelet\@let@token\@onedot}
\def\@onedot{\ifx\@let@token.\else.\null\fi\xspace}

\makeatother

\definecolor{MyDarkBlue}{rgb}{0,0.08,0.8}
\definecolor{MyDarkGreen}{RGB}{45,155,45}
\definecolor{MyDarkRed}{rgb}{0.8,0.02,0.02}
\definecolor{MyOrange}{rgb}{1.0, 0.4, 0.2}
\definecolor{MyPurple}{RGB}{111,0,255}
\definecolor{MyRed}{rgb}{0.8,0.0,0.0}
\definecolor{MyGold}{rgb}{0.75,0.6,0.12}
\definecolor{MyDarkgray}{rgb}{0.66, 0.66, 0.66}
\definecolor{MyDeepGreen}{rgb}{0.0, 0.4, 0.3}
\definecolor{MyPlumGray}{rgb}{0.33, 0.28, 0.39}
\definecolor{JiayuanColor}{rgb}{0.60,0.43,0.48}

\usepackage{hyperref}

\usepackage{orcidlink}

\begin{document}

\title{Layering Virtual Try-On} 



\author{Chun Feng \and Bowei Chen \and
Mengyi Shan \and Ira Kemelmacher-Shlizerman}

\authorrunning{C.~Feng et al.}


\institute{University of Washington, Seattle WA 98195, USA \\ \email{\{chunfeng, boweiche, shanmy, kemelmi\}@cs.washington.edu}}

\maketitle

\begin{figure}
    \centering
    \includegraphics[width=1.0\linewidth]{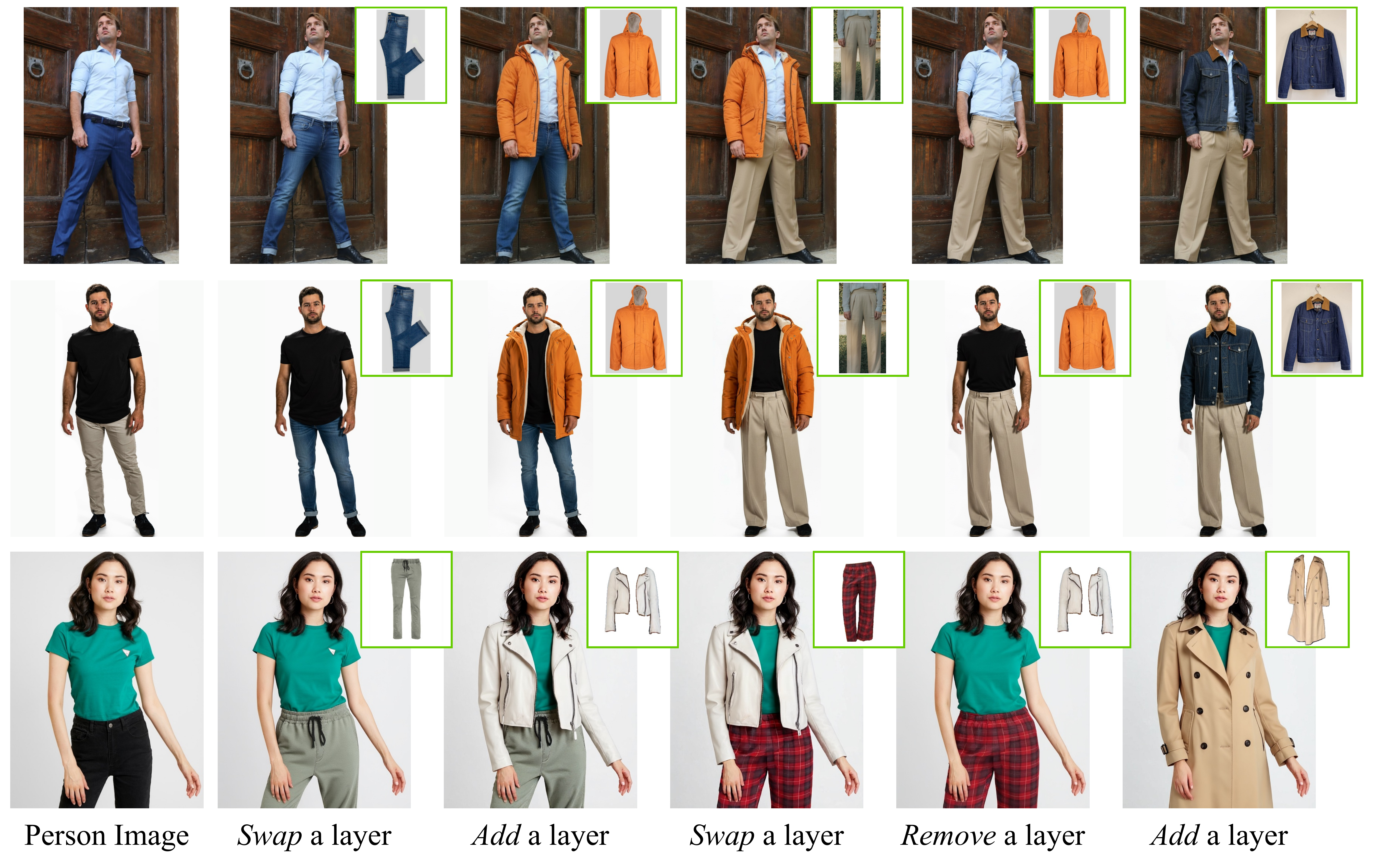}
    \vspace{-0.75cm}
    \caption{
        We propose a Layering Virtual Try-On (LVTON) method that allows for swapping, adding, and removing clothing layers on any given person image.
        Given a person image, a garment image, and a textual instruction (e.g., swap, add, or remove), it generates a try-on image reflecting the desired garment edit while preserving inner layers when necessary. We showcase three sequential editing examples that exhibit natural layering, accurate garments, consistent facial identity, and realistic shading within complex backgrounds. 
    }
    \label{fig:teaser}
\end{figure}

\vspace{-1.2cm}

\begin{abstract}
In the real world, fashion is about layering: adding a jacket over a shirt, or a sequence of adding and removing layers, rather than just a single-layer swap. This fundamental real-world task remains a challenge in existing Virtual Try-On (VTON) methods, which excel at single-layer replacement but are not designed to layer or de-layer an existing outfit. This paper proposes Layering Virtual Try-On (LVTON), a layering benchmark and method that preserves an existing outfit while enabling sequential layering. We find that current VTON paradigms are fundamentally ill-equipped for LVTON, as their reliance on cloth-agnostic representations and single-item datasets discards essential layering context. Our key insight is that the LVTON challenge must be disentangled into two distinct competencies: (1) General VTON Priors (e.g., deformation, identity preservation) and (2) Specific Layering Knowledge (e.g., layering order and occlusion reasoning). First, our model obtains general VTON priors by being trained on data produced by an automatic data generation pipeline that synthesizes samples from fashion videos via segmentation and inpainting. Second, the model is fine-tuned on a small, dedicated LVTON dataset to learn the layering logic. Our method achieves state-of-the-art results on our LVTON benchmark and demonstrates superior generalizability on traditional VTON benchmarks, setting new state-of-the-art results when fine-tuned and exhibiting zero-shot capabilities.

\keywords{Image Diffusion Models \and Image Edit \and Virtual Try-On}
\end{abstract}

\vspace{-1.0cm}

\begin{figure}[h]
    \centering
    \includegraphics[width=1.0\textwidth]{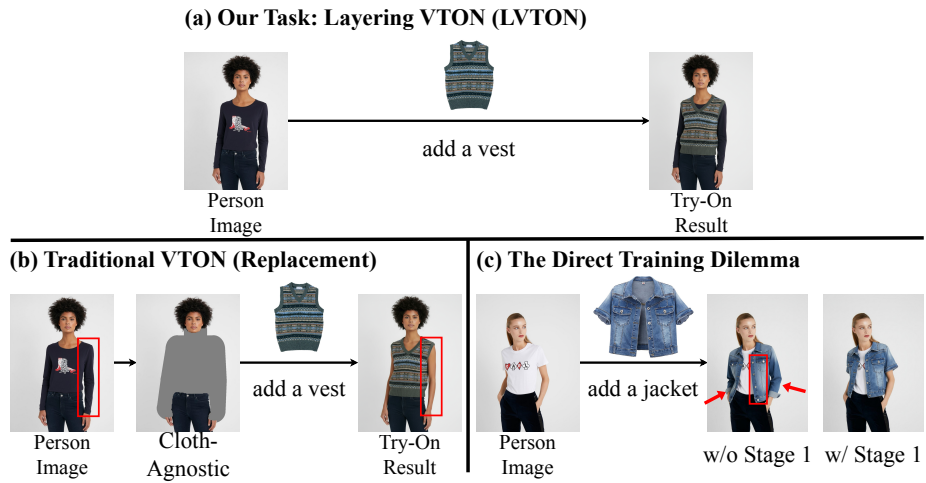}
    \vspace{-0.7cm}
    \caption{Illustration of the Layering VTON (LVTON) challenge and the motivation for our two-stage paradigm. 
    (a) LVTON task aims to composite a new garment (e.g., a sweater vest) over an existing outfit while preserving inner layers.
    (b) Traditional VTON paradigms mainly rely on a cloth-agnostic representation by masking out the original garment, which discards the inner layer. (c) The direct training dilemma: a model trained only on layering data (w/o Stage 1) learns correct composition logic (preserving the shirt) but fails to retain garment identity (e.g., jacket buttons and sleeves). Our two-stage approach (w/ Stage 1) leverages general VTON priors to resolve this issue.
    }
    \label{fig:intro_motivation}
    \vspace{-0.8cm}
\end{figure}

\section{Introduction}
\label{sec:intro}

Virtual Try-On (VTON) has emerged as a transformative technology in computer vision, with the potential to revolutionize e-commerce, digital fashion, and augmented reality. In its canonical form, the task involves rendering a target garment onto an image of a person to synthesize a photorealistic result. Driven by advances in diffusion models and generative adversarial networks, state-of-the-art (SOTA) methods~\cite{choi2021viton, lee2022high, xie2023gp, morelli2023ladi, gou2023taming, kim2024stableviton, xu2025ootdiffusion, choi2024improving, jiang2024fitdit, chong2024catvton, feng2025omnitry, guo2025any2anytryon, deria2025muga, chen2024wear, velioglu2025enhancing, sun2024outfitanyone, zhu2024m, zhang2024mmtryon} have achieved remarkable fidelity in single-garment replacement. 

Despite impressive progress, existing approaches remain constrained to simplified single-garment scenarios, limiting their applicability in real-world dressing contexts. Human apparel is inherently layered; individuals wear multiple garments that occlude and interact with one another: a shirt collar protruding from a sweater, or an inner t-shirt visible beneath an open jacket. The capability to compose a new garment onto an existing outfit rather than merely replacing it, represents a critical yet overlooked frontier for VTON's practical deployment. In this work, we formally define this task as \textbf{Layering Virtual Try-On (LVTON)}. Unlike conventional VTON, which replaces the existing garment, LVTON aims to composite a new garment on top of the person's existing outfit, while preserving the visibility of the inner layers, as illustrated in~\cref{fig:intro_motivation} (a).

Existing VTON paradigms are fundamentally ill-equipped to address LVTON. We identify two root causes. First, a methodological limitation prevalent in many dominant approaches: a cloth-agnostic person representation is needed~\cite{choi2021viton, lee2022high, xie2023gp, morelli2023ladi, gou2023taming, kim2024stableviton, xu2025ootdiffusion, choi2024improving, jiang2024fitdit, deria2025muga, chen2024wear, velioglu2025enhancing, sun2024outfitanyone, zhu2024m}. This process, which masks out the original garment to simplify the single-replacement task, irreversibly discards the essential contextual information --- the very outfit upon which a new layer must be added. Second, a more pervasive systemic data bias: popular benchmarks, such as VITON-HD~\cite{choi2021viton} and DressCode~\cite{morelli2022dress}, consist exclusively of single-item try-on examples. Models trained on this data have never observed the complex interplay of layered garments and thus cannot learn the required composition logic. As shown in~\cref{fig:intro_motivation} (b), previous methods consequently fail on the LVTON task, often erasing the inner layer or producing severe artifacts. 

The failure of these mask-based approaches necessitates a paradigm shift towards mask-free editing. Large-scale pre-trained diffusion models~\cite{wu2025qwen}, which excel at high-fidelity, text-guided image editing without relying on explicit masks, thus emerge as a natural starting point for LVTON. A straightforward solution would be to fine-tune such a model on a dedicated LVTON dataset. Yet, this approach faces two critical hurdles inherent to real-world data collection. First is extreme data scarcity: collecting large-scale, high-quality layering pairs with diverse garments and subjects is prohibitively expensive and practically infeasible. Therefore, any viable \textbf{LVTON solution must be highly data-efficient}. Second~is the inevitable pose variance. Unlike standard VTON datasets with perfectly aligned pairs, it is practically impossible for a subject to maintain an identical pose while donning an additional layer. This introduces spatial misalignment, significantly compounding the task difficulty compared to standard fixed-pose VTON. Our preliminary experiments reveal that fine-tuning on such limited and pose-variant data is a critical flaw. The model is forced to learn complex layering logic, spatial alignment, and fundamental VTON capabilities (e.g., texture fidelity, garment deformation) simultaneously from an insufficient number of examples. As illustrated in~\cref{fig:intro_motivation} (c), this leads to a loss of fidelity, where the model fails to maintain the garment's identity (e.g., buttons) and struggles with realistic deformation, even as it learns the correct layering logic (e.g., preserving the inner shirt collar).

This leads to our key insight: the challenges of LVTON are two-fold and must be disentangled. We posit that LVTON requires (1) general VTON priors (i.e., robust garment deformation, placement, and identity preservation) and (2) specific layering knowledge (i.e., occlusion reasoning and composition logic). The failure of direct training stems from forcing a model to learn both complex knowledge simultaneously from scarce LVTON data. We argue that general VTON priors and specific layering knowledge are distinct yet complementary. By disentangling them, we can exploit abundant single-layer data to learn robust priors, while using only limited layering data for specialization. From this, we propose a two-stage disentangled training paradigm. 

\vspace{-0.3cm}
\paragraph{\textbf{Stage 1: Learning general VTON priors.}} 
The goal of this stage is to equip the model with robust VTON priors without relying on scarce LVTON data. However, standard VTON datasets are ill-suited for this purpose due to the dual challenges identified above: (i) their reliance on cloth-agnostic masking discards the context vital for layering, and (ii) their aligned poses fail to provide the spatial deformation supervision needed to handle the pose variance inherent in LVTON. If the model does not learn to handle pose shifts in this stage, it cannot effectively support the layering task in stage 2. To address this, we introduce an automatic data generation pipeline that leverages ubiquitous fashion videos. This approach produces training pairs that are, by design, both mask-free and pose-variant. By learning from video frames with natural body movements, the model is forced to master complex garment deformation and identity preservation, thereby establishing a powerful prior robust to the spatial misalignment encountered in real-world layering.

\vspace{-0.3cm}
\paragraph{\textbf{Stage 2: Learning the specific layering knowledge.}}
With the general VTON priors from stage 1, we now focus on learning the specific layering knowledge by fine-tuning on our curated LVTON dataset. We acknowledge that building a comprehensive, unbiased layering dataset at scale remains a profound challenge. However, our disentangled approach turns this limitation into a demonstration of data efficiency. Because the model is already an expert in general VTON from stage 1, it does not need to relearn fundamental garment deformation or texture fidelity. Instead, it can dedicate its capacity to quickly learning the nuanced compositional logic of layering from this scarce data.

Our extensive experiments validate the effectiveness of our disentangled paradigm. On our newly collected LVTON benchmark, our method surpasses existing VTON baselines, producing photorealistic LVTON results where prior art fails. Furthermore, general VTON priors learned in stage 1 prove highly transferable; it exhibits strong zero-shot performance on standard VTON benchmarks. With standard fine-tuning, our model achieves state-of-the-art results on these traditional tasks, demonstrating the robustness and generality of our approach. Finally, our model achieves sequential outfit editing, as shown in~\cref{fig:teaser}.

Our contributions are summarized as follows:
\vspace{-0.2cm}
\begin{itemize}
    \setlength\itemsep{-0em}
    \item We define and address Layering Virtual Try-On, a novel and critical task for real-world VTON application.
    \item We propose a highly data-efficient, disentangled training paradigm that decomposes the LVTON task, enabling the model to learn complex layering logic from a strictly limited dataset and to generalize to in-the-wild scenarios.
    \item We introduce an automatic and scalable data generation pipeline from videos, enabling the model to learn general VTON priors without requiring real-world paired data.
    \item We experimentally validate our paradigm, achieving SOTA on our LVTON benchmark and demonstrating competitive performance on traditional VTON benchmarks.
\end{itemize}

\section{Related Works} 
\label{sec:related_works}

\paragraph{\textbf{Virtual Try-On systems.}} 
Virtual Try-On (VTON) has seen significant progress, with the majority of existing methods \cite{choi2021viton, lee2022high, xie2023gp, morelli2023ladi, gou2023taming, kim2024stableviton, xu2025ootdiffusion, choi2024improving, jiang2024fitdit, deria2025muga, chen2024wear, velioglu2025enhancing, sun2024outfitanyone, zhu2024m, zhang2024mmtryon} operating under a cloth-agnostic setting. Early GAN-based works \cite{choi2021viton, lee2022high, xie2023gp} typically employ a two-stage process involving garment warping and try-on generation, both of which heavily rely on precise masks to define the synthesis region. More recent diffusion-based approaches \cite{morelli2023ladi, gou2023taming, kim2024stableviton, xu2025ootdiffusion, deria2025muga, chen2024wear, velioglu2025enhancing, sun2024outfitanyone, zhu2024m, zhang2024mmtryon} merge these stages into an end-to-end process, but still require the cloth-agnostic representation as a conditional input. As discussed in~\cref{sec:intro}, this fundamental reliance on masks precludes the modeling of layering try-on, as all information about the person's existing outfit is erased.

Acknowledging this limitation, a few recent works have proposed mask-free VTON \cite{chong2024catvton, feng2025omnitry, guo2025any2anytryon, zhang2024mmtryon}. These methods synthesize the try-on result directly on the original, fully-clothed person image, for example, by utilizing different conditioning strategies \cite{chong2024catvton, guo2025any2anytryon, zhang2024mmtryon} or leveraging powerful backbones \cite{feng2025omnitry}. However, while this removes the mask dependency, it remains insufficient for true layering capabilities. The core issue persists: these mask-free methods are still trained on traditional VTON datasets \cite{choi2021viton, morelli2022dress} or custom-collected benchmarks \cite{feng2025omnitry} that lack the necessary compositional examples of a person sequentially adding garments. These benchmarks predominantly contain single-item replacement pairs rather than the sequential ``before-and-after'' examples needed for layering. Therefore, while our method operates in the mask-free setting, our primary contribution is not the mask-free paradigm itself. Instead, we address the critical data bottleneck that prevents existing mask-free methods from achieving realistic layering.


\vspace{-0.3cm}
\paragraph{\textbf{Generative data synthesis for training.}} 
Deep generative models have increasingly been employed to synthesize paired training data, circumventing the high cost of manual collection \cite{eigenschink2023deep, yang2020generative, wang2023self, honovich2023unnatural, tian2023stablerep, fan2024scaling, josifoski2023exploiting, bonifacio2022inpars, hammoud2024synthclip, liu2020data, ng2020ssmba}. A common strategy involves leveraging inpainting models to alter garments on a single image, thereby creating a pseudo-pair \cite{yin2023ttida, shivashankar2023semantic, guo2025any2anytryon, feng2025omnitry, valvano2024controllable, fu2024dreamda, li2025real, li2025enhancing, wang2025text2data, huang2025can}. However, a critical limitation of these approaches is that the synthesized input and target are pose-invariant. While acceptable for standard VTON, this lack of pose variation is fundamentally insufficient for LVTON. As established in~\cref{sec:intro}, a model trained solely on pose-invariant pairs learns trivial pixel-to-pixel correspondence rather than the complex spatial deformation required to align a garment to a shifted pose. To address this, our stage 1 introduces a cross-frame data generation paradigm that mines pose-variant pairs from unstructured videos. By pairing an inpainted frame with a temporally distant frame from the same video, we construct training samples where the subject's pose differs naturally. This forces the model to learn robust garment deformation and alignment priors, which are indispensable for handling the spatial misalignment in LVTON. 
Crucially, by mastering complex spatial and deformation priors, we reduce the burden on subsequent training phases. This directly enables the high sample efficiency of our stage 2, allowing the model to learn specific layering logic from a strictly limited dataset.

\section{Methods}
\label{sec:methods}

Our method implements the two-stage disentangled training paradigm, designed to address the data and methodological challenges of VTON for layered apparel.
We first detail the two core stages of our paradigm: (1) an automatic data generation pipeline to learn the general VTON priors (\cref{sec:stage1}), and (2) a specialized fine-tuning strategy using an augmented, real-world dataset to learn the specific layering knowledge (\cref{sec:stage2}). We then describe our model architecture and model training (\cref{sec:model}).

\vspace{-0.3cm}
\paragraph{\textbf{Task Definition.}}
Given a person image $\mathcal{I}_{M}$, a garment image $\mathcal{I}_{G}$, a target pose $\mathcal{I}_{P}$, and a text description $\mathcal{T}$, our goal is to generate a realistic image $\mathcal{I}_{T}$ that reflects the desired garment and pose edit while preserving the person's identity and the exact visual integrity of non-target (inner) clothing layers. Standard VTON aims to swap an entire garment. In contrast, the text prompt $\mathcal{T}$ specifies the intended interaction (e.g., ``add a dark blue denim blouse,'' ``remove a gray sweater,'' or ``swap X for Y''), requiring the model to modify only a specified clothing region while preserving the non-target (inner) regions from $\mathcal{I}_{M}$.


\begin{figure}[t]
    \centering
    \includegraphics[width=1.0\linewidth]{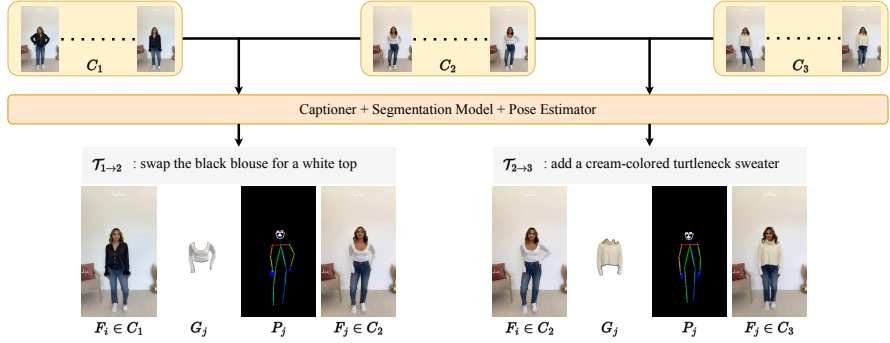}
    \vspace{-0.7cm}
    \caption{Our two-stage pipeline. (a) The first stage builds general VTON priors by synthesizing mask-free, pose-mismatched training pairs from videos. This forces the model to learn robust spatial deformation, ensuring accurate alignment even during standard fixed-pose inference. (b) The second stage uses a small, real-world layering dataset to teach the model specific layering knowledge (e.g., ``add'').}
    \label{fig:pipeline}
    \vspace{-0.7cm}
\end{figure}


\subsection{Stage 1: Learning the General VTON Priors}
\label{sec:stage1}

As discussed in~\cref{sec:intro}, the core challenge of LVTON lies in the incompatibility of standard mask-based VTON methods, which discard essential contextual cues. To overcome this incompatibility, we first learn general VTON priors in a mask-free manner. This stage models garment deformation, shading, and spatial placement, providing a strong foundation for context-aware editing.

Our approach is built on a key observation: abundant online videos feature people wearing the same outfits in diverse poses. We utilize this property to generate a scalable, challenging synthetic dataset. The process involves four steps (illustrated in~\cref{fig:pipeline}, (a)): \textbf{(1)} Data Collection: we collect video sequences of subjects in consistent outfits but with varying poses. \textbf{(2)} Frame Extraction: we extract frames and process them using \cite{ravi2024sam, yang2023effective, bai2025qwen2} to obtain tuples $\{ (F_{i}, M_{i}, G_{i}, P_{i}, \mathcal{T}_{i}) \}_{i = 1}^{N}$, where $F_{i}$ is the image, $M_{i}$ the garment mask, $G_{i}$ the segmented garment, $P_{i}$ the pose, and $\mathcal{T}_{i}$ is the corresponding text description of the garment $G_i$. \textbf{(3)} Synthesis: we employ an inpainting model \cite{batifol2025flux} to synthesize a new version of each frame, $F_{i}^{\prime}$, featuring a novel outfit guided by a Large Language Model (LLM) generated prompt. \textbf{(4)} Training Pair Construction: we construct the final training pairs by sampling distinct frames $F_{i}$ and $F_{j}$ from the same sequence ($i \neq j$). Each training instance is constructed as $(\mathcal{I}_{M}, \mathcal{I}_{G}, \mathcal{I}_{P}, \mathcal{T}, \mathcal{I}_{T}) := (F_{i}^{\prime}, G_{j}, P_{j}, \mathcal{T}_{j}, F_{j})$. More details can be found in Appendix B.1. 

This video-based construction is pivotal for disentangling the learning process. Crucially, the objective of stage 1 is \textbf{NOT} to approximate LVTON, but rather to isolate the fundamental mechanics of mask-free editing. First and foremost, the task forces the model to reason about garment placement directly from context without masks. More critically, by enforcing $i \neq j$, our pipeline guarantees that the input pose ($P_{i}$ in $F_{i}'$) and target pose ($P_{j}$) are spatially misaligned. As emphasized in~\cref{sec:intro}, real-world layering inevitably introduces pose variance. If the model were trained on pose-invariant pairs (where $i=j$), it would collapse into learning trivial pixel alignment. By solving the harder task of ``deform-and-place'' in this stage, the model acquires robust spatial priors that serve as the indispensable structural foundation for the delicate layering composition in stage~2. Finally, sourcing from ``in-the-wild'' videos exposes the model to diverse lighting and backgrounds, preventing overfitting to sterile studio environments.


\begin{figure}[t]
    \centering
    \includegraphics[width=\textwidth]{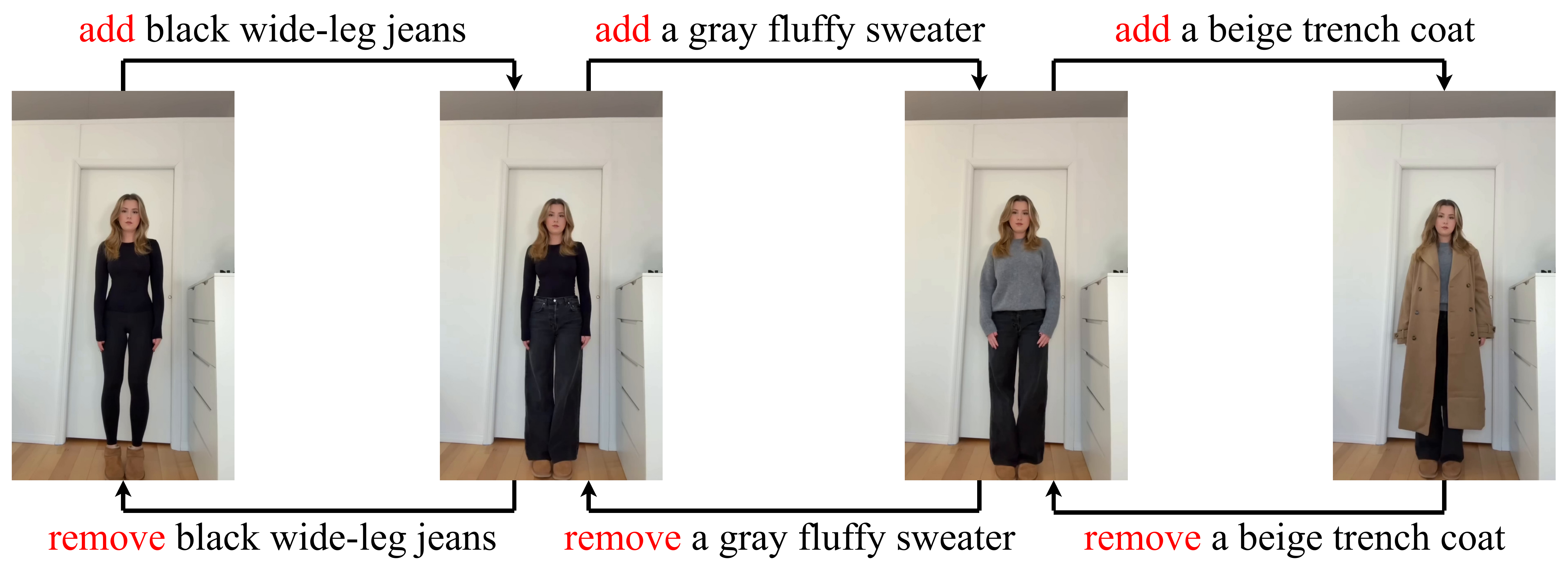}
    \vspace{-0.7cm}
    \caption{Visualizations of fine-tuning data. \textbf{Top}: A real-world ``add'' sequence collected from video. \textbf{Bottom}: The corresponding ``remove'' sequence generated by temporal reversal augmentation.}
    \label{fig:stage2_and_model}
    \vspace{-0.6cm}
\end{figure}

\subsection{Stage 2: Fine-tuning for Layering Knowledge}
\label{sec:stage2}

While stage 1 equips the model with strong general VTON priors, it lacks supervision over explicit compositional operations (e.g., ``add''). We therefore fine-tune the model on a real-world dataset designed to teach such compositional reasoning. As noted in~\cref{sec:intro}, building a layering dataset at scale is infeasible. Our disentangled paradigm turns this limitation into a demonstration of data efficiency: based on general VTON priors learned in stage 1, the model can dedicate its entire learning capacity to mastering composition logic. We propose two contributions: a data collection pipeline and a novel augmentation technique to expand the dataset effectively. More details can be found in Appendix B.2.

\vspace{-0.3cm}
\paragraph{\textbf{Data collection.}}
We collect videos of individuals sequentially adding or swapping garments. We partition video frames into temporal clusters $\{C_{1}, C_{2}, \cdots, C_{K}\}$ based on stable outfit configurations. We employ a Vision-Language Model \cite{bai2025qwen2} to automatically generate textual descriptions $\mathcal{T}_{m \rightarrow m + 1}$ for each transition (e.g., ``add a gray sweater''). We then sample a source frame $F_{i} \in C_{m}$ and a target frame $F_{j} \in C_{m+1}$ from adjacent clusters to form the training tuple as $(\mathcal{I}_{M}, \mathcal{I}_{G}, \mathcal{I}_{P}, \mathcal{T}, \mathcal{I}_{T}) := (F_{i}, G_{j}, P_{j}, \mathcal{T}_{m \rightarrow m + 1}, F_{j})$. 

\vspace{-0.3cm}
\paragraph{\textbf{Temporal reversal augmentation.}}
To mitigate the scarcity of our fine-tuning data, we introduce a temporal reversal augmentation technique. This stems from the observation that a ``dressing'' sequence, when reversed, becomes a valid ``undressing'' sequence. For each forward training instance $(F_{i}, G_{j}, P_{j}, \mathcal{T}_{m \rightarrow m + 1}, F_{j})$, we create a corresponding reverse instance $(F_{j}, G_{i}, P_{i}, \mathcal{T}_{m+1 \rightarrow m}, F_{i})$, where the reverse text prompt $\mathcal{T}_{m+1 \rightarrow m}$ is generated by applying inversion rules (e.g., ``add \texttt{<garment>}'' becomes ``remove \texttt{<garment>}'') (\cref{fig:stage2_and_model}). This effectively doubles our real-world data and enables learning both garment addition and removal.


\begin{figure}[h]
    \centering
    \includegraphics[width=1.0\linewidth]{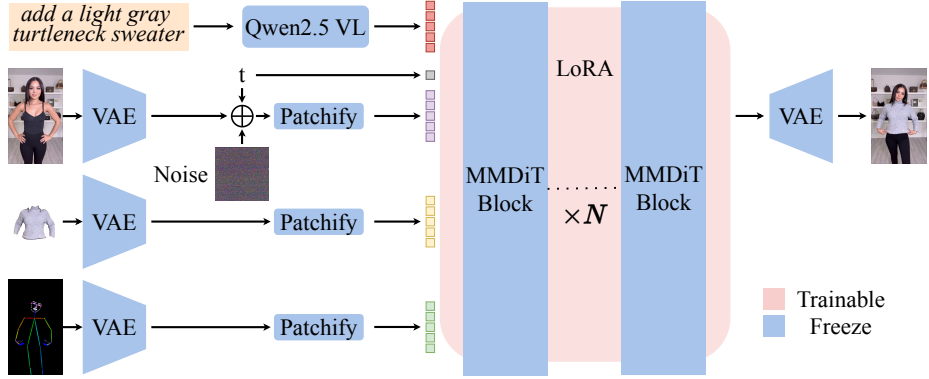}
    \vspace{-0.5cm}
    \caption{An overview of model architecture. We apply Low-Rank Adaptation (LoRA) to fine-tune Qwen-Image-Edit. The model conditions on four inputs: (1) a text prompt (e.g., ``add a light gray turtleneck sweater''), processed by the Qwen2.5-VL; (2) a noised person image latent, created by adding noise at timestep $t$; (3) a garment image latent; and (4) a pose image latent. }
    \label{fig:architecture}
    \vspace{-0.7cm}
\end{figure}

\subsection{Model Architecture and Training}
\label{sec:model}

\paragraph{\textbf{Model architecture.}}
We build on the pre-trained Qwen-Image-Edit \cite{wu2025qwen}, a model with strong high-fidelity image editing capability. As shown in~\cref{fig:architecture}, it comprises three main components: (1) a Variational AutoEncoder (VAE) as the image tokenizer, (2) Qwen2.5-VL \cite{bai2025qwen2} as the multimodal condition encoder, and (3) a Multimodal Diffusion Transformer (MMDiT) \cite{esser2024scaling} as the diffusion backbone. Its key innovation is a dual-stream design, where Qwen2.5-VL encodes high-level semantics from text and reference images, while the VAE provides low-level visual latents. The MMDiT fuses both streams, achieving a balance between semantic consistency and visual fidelity. 

\vspace{-0.3cm}
\paragraph{\textbf{Model training.}}
We train the model using LoRA \cite{hu2022lora} via our two-stage training paradigm. We first train it on our synthetic dataset (stage 1) and then fine-tune it on the real-world layering dataset (stage 2). The model architecture (\cref{sec:model}), training objective (standard denoising loss), and all optimization details are kept identical across stage 1 and stage 2. More details can be found in Appendix A.


\vspace{-0.55cm}

\begin{table}[h]
    \small
    \centering
    \caption{
        Quantitative comparisons on our LVTON dataset. Our method performs best across all metrics. Note: absolute FID scores are inflated due to our small test set ($532$ images). Best results are in \textbf{bold}. $\uparrow$ indicates higher is better, $\downarrow$ indicates lower is better.
    }
    \vspace{-0.3cm}
    \begin{tabular}{l c@{\hspace{0.2cm}}c cccc}
        \toprule
         Model & && SSIM $\uparrow$ & LPIPS $\downarrow$ & FID $\downarrow$ & KID{\tiny $\times 10^{3}$} $\downarrow$ \\
        
        \midrule

        
        OmniTry \cite{feng2025omnitry}          & && $0.810$  & $0.176$ & $54.585$ & $3.165$ \\
        
        Any2AnyTryon \cite{guo2025any2anytryon} & && $0.787$ & $0.229$ & $89.567$ & $18.645$ \\
        
        Nano Banana \cite{comanici2025gemini}   & && $0.798$ & $0.195$ & $49.179$ & $2.362$ \\

        HunyuanImage-3.0-Instruct \cite{cao2025hunyuanimage}   & && $0.767$ & $0.246$ & $88.070$ & $15.561$ \\
        
        Ours                                    & && $\mathbf{0.843}$ & $\mathbf{0.127}$ & $\mathbf{48.957}$ & $\mathbf{1.463}$ \\
        
        \bottomrule
    \end{tabular}
    \label{table:layering_result_table}
    \vspace{-0.6cm}
\end{table}

\begin{figure}[t]
    \centering
    \includegraphics[width=1.0\linewidth]{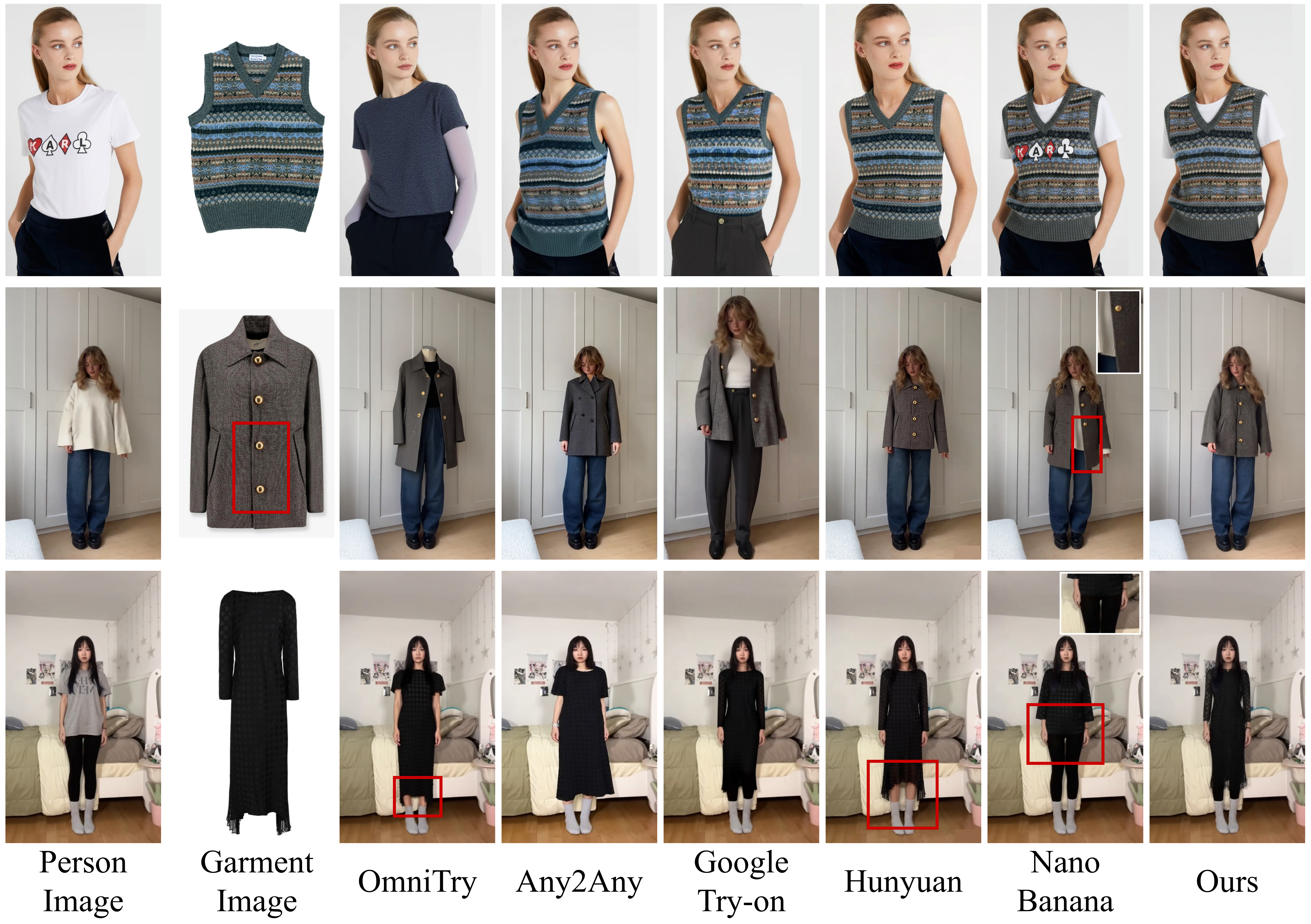}
    \vspace{-0.7cm}
    \caption{
        Qualitative results of LVTON. Our method successfully composites the target garment while preserving the inner layer. Baselines often erase inner details or introduce artifacts. Regions corresponding to the errors made by the baselines are highlighted with red boxes in their respective predictions for comparison (e.g., Nano Banana fails to preserve the lower portion of the black dress).
    }
    \label{fig:layering_result_figure}
    \vspace{-0.6cm}
\end{figure}


\section{Experiments}
\label{sec:experiments}

In this section, \cref{sec:layering_vton} and \cref{sec:classical_vton} evaluate the proposed method on Layering VTON and traditional VTON tasks, respectively; \cref{sec:user_study} provides user study and layering-specific evaluation; \cref{sec:in_the_wild} shows in-the-wild results; and \cref{sec:ablation} presents ablation studies to validate our key design choices.


\subsection{Layering Virtual Try-On}
\label{sec:layering_vton}

\paragraph{\textbf{Datasets and evaluation metrics.}} 
Our training paradigm employs two distinct stages to address the scarcity of real-world layering data. For stage 1 (general VTON priors), we construct a synthetic dataset of $29,151$ training pairs from $362$ videos (\cref{sec:stage1}). For stage 2 (layering knowledge), we curate a real-world LVTON dataset of $5,768$ training pairs and $532$ test pairs from $60$ videos, expanded via temporal reversal augmentation (\cref{sec:stage2}). We evaluate using SSIM~\cite{wang2004image}, LPIPS~\cite{zhang2018unreasonable}, FID~\cite{heusel2017gans}, and KID~\cite{binkowski2018demystifying} (note that absolute FID values are inflated due to the small test set). More details can be found in Appendix C.

\vspace{-0.3cm}
\paragraph{\textbf{Baselines and results.}} 
We conduct a comparison with mask-free methods, OmniTry \cite{feng2025omnitry}, and Any2AnyTryon \cite{guo2025any2anytryon}, as well as three industrial models: HunyuanImage-3.0-Instruct~\cite{cao2025hunyuanimage}, Nano Banana \cite{comanici2025gemini}, and Google Tryon (a variant of \cite{baldridge2024imagen}), on our LVTON dataset. To ensure a fair comparison, we adapt these baselines by providing layering instructions (e.g., ``add a shirt'') as text prompts. For Nano Banana, evaluation is conducted via its public API. The Google TryOn comparison is limited to qualitative results because it lacks a public API. HunyuanImage-3.0-Instruct is evaluated via its official open-source implementation. Quantitative and qualitative results are shown in~\cref{table:layering_result_table} and \cref{fig:layering_result_figure}, respectively. Our model outperforms all baselines across metrics, setting a new SOTA for LVTON.


\begin{table}[t]
    \scriptsize
    \centering
    \caption{
        Quantitative comparison with SOTA methods on the VITON-HD~\cite{choi2021viton} and DressCode~\cite{morelli2022dress} benchmarks. Ours is our complete two-stage model, fine-tuned on the benchmarks. Stage 1-only is our stage 1 model, evaluated in a zero-shot setting (\cref{sec:ablation}). Best results are in \textbf{bold}.
    }
    \vspace{-0.3cm}
    \begin{tabular}{l c@{\hspace{-0.1cm}}c cccc c@{\hspace{0.1cm}}c cccc}
        \toprule
        \multirow{2}{*}{Model} && & \multicolumn{4}{c}{VITON-HD} && & \multicolumn{4}{c}{DressCode} \\
  && & SSIM $\uparrow$ & LPIPS $\downarrow$ & FID $\downarrow$ & KID{\tiny $\times 10^{3}$} $\downarrow$ && & SSIM $\uparrow$ & LPIPS $\downarrow$ & FID $\downarrow$ & KID{\tiny $\times 10^{3}$} $\downarrow$ \\
        \midrule
        \multicolumn{13}{c}{\textbf{Methods Using Cloth-Agnostic Representation}} \\
        \midrule
        VITON-HD \cite{choi2021viton} &&          & $0.862$ & $0.117$ & $12.117$ & $3.230$  && & -       & -       & -         & -       \\
        HR-VITON \cite{lee2022high} &&            & $0.878$ & $0.105$ & $11.270$ & $2.730$  && & $0.936$ & $0.065$ & $13.820$  & $2.710$ \\
        GP-VTON \cite{xie2023gp} &&               & $0.894$ & $0.079$ & $9.197$  & $0.880$  && & $0.948$ & $0.268$ & $11.980$  & $8.170$ \\
        LADI-VTON \cite{morelli2023ladi} &&
                                           & $0.876$ & $0.091$ & $9.410$  & $1.600$ &&          & $0.915$ & $0.063$          & $13.260$ & $2.670$ \\
        DCI-VTON \cite{gou2023taming} &&          & $0.880$ & $0.081$ & $8.754$  & $1.100$ &&          & $0.937$ & $\mathbf{0.031}$ & $10.820$ & $1.890$ \\ 
        StableVITON \cite{kim2024stableviton} &&  & $0.852$ & $0.084$ & $8.698$  & $0.880$ &&          & $0.911$ & $0.050$          & $11.266$ & $0.720$ \\
        OOTDiffusion \cite{xu2025ootdiffusion} && & $0.878$ & $0.071$ & $8.810$  & $0.820$ &&          & $0.898$ & $0.073$          & $3.950$  & $0.720$ \\
        IDM-VTON \cite{choi2024improving} &&      & $0.870$ & $0.102$ & $6.290$  & $1.201$ &&          & $0.920$ & $0.062$          & $8.640$  & $2.904$ \\
        FitDit \cite{jiang2024fitdit} &&          & $0.899$ & $0.066$ & $4.731$  & $\mathbf{0.190}$ && & $0.926$ & $0.043$          & $2.638$  & $0.499$ \\
        \midrule
        \multicolumn{13}{c}{\textbf{Mask-Free Methods}} \\
        \midrule
        CATVTON \cite{chong2024catvton} &&         & $0.870$ & $0.057$ & $5.425$  & $0.411$ && & $0.892$ & $0.046$ & $3.992$  & $0.818$ \\
        OmniTry \cite{feng2025omnitry} &&          & $0.698$ & $0.368$ & $57.871$ & $0.450$ && & $0.842$ & $0.188$ & $25.364$ & $0.410$ \\
        Any2AnyTryon \cite{guo2025any2anytryon} && & $0.839$ & $0.088$ & $6.934$ & $0.740$ && & $0.889$ & $0.138$ & $15.719$ & $0.502$ \\ 
        MuGa-VTON \cite{deria2025muga} && & $0.898$ & $0.073$ & $7.240$ & $0.520$ && & $0.937$ & $0.046$ & $8.930$ & $0.940$ \\
        \midrule
        Stage 1-only && & $0.887$          & $0.086$          & $5.235$          & $1.550$ && & $0.898$          & $0.101$ & $4.710$          & $0.890$ \\
        Ours  &&         & $\mathbf{0.928}$ & $\mathbf{0.055}$ & $\mathbf{3.235}$ & $0.200$ && & $\mathbf{0.948}$ & $0.042$ & $\mathbf{2.598}$ & $\mathbf{0.315}$ \\
        \bottomrule
    \end{tabular}
    \label{table:classical_result_table}
    \vspace{-0.3cm}
\end{table}


\begin{figure}[h]
    \centering
    \includegraphics[width=1.0\linewidth]{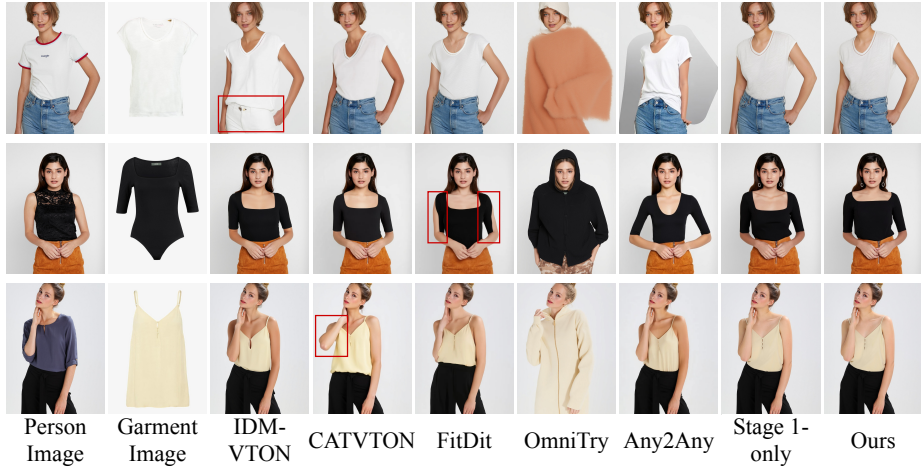}
    \vspace{-0.7cm}
    \caption{
        Qualitative comparison on the VITON-HD~\cite{choi2021viton} and DressCode~\cite{morelli2022dress} benchmarks. Ours represents our complete two-stage model. Stage 1-only represents our stage 1 model evaluated zero-shot (analysis in~\cref{sec:ablation}). Red boxes highlight artifacts and failure cases in the baseline results. Ours consistently avoids these issues, demonstrating superior fidelity and generalization.
    }
    \label{fig:classical_result_figure}
    \vspace{-0.65cm}
\end{figure}

\subsection{Traditional Virtual Try-On}
\label{sec:classical_vton}

\paragraph{\textbf{Datasets and evaluation metrics.}}
We also evaluate our model on traditional VTON benchmarks (VITON-HD \cite{choi2021viton}, DressCode \cite{morelli2022dress}) using the same metrics.

\vspace{-0.3cm}
\paragraph{\textbf{Baselines and results.}} 
We benchmark our full two-stage model, Ours, against SOTA methods. As our model requires pose and text inputs, we use the same VLM and pose estimator \cite{bai2025qwen2, yang2023effective} to generate them for the benchmarks. As shown in~\cref{table:classical_result_table}, Ours outperforms all existing methods in SSIM and FID, setting a new state of the art on these benchmarks. 
The corresponding qualitative results in~\cref{fig:classical_result_figure} also showcase our model's superior fidelity and generalization.


\vspace{-0.6cm}
\begin{table}[h]
    \centering
    \caption{
        User study results ($5$-point Likert scale). Our method outperforms baselines across both general generation quality (i.e., Realism, Instr.) and LVTON-specific layering metrics (i.e., Occlusion, Inner-Vis., Overall-Lay.).
    }
    \vspace{-0.3cm}
    \begin{tabular}{lccccc}
        \hline
        Method & Realism & Instr. & Occlusion & Inner-Vis. & Overall-Lay. \\
        \hline
        OmniTry \cite{feng2025omnitry} & $3.370$ & $3.521$ & $3.603$ & $3.125$ & $2.945$ \\
        Any2Any \cite{guo2025any2anytryon} & $3.194$ & $3.069$ & $3.639$ & $2.859$ & $2.761$ \\
        Nano Banana \cite{comanici2025gemini} & $4.278$ & $2.361$ & $3.486$ & $3.394$ & $2.833$ \\
        Ours & $\mathbf{4.597}$ & $\mathbf{4.694}$ & $\mathbf{4.750}$ & $\mathbf{4.535}$ & $\mathbf{4.569}$ \\
        \hline
    \end{tabular}
    \label{table:user_study}
    \vspace{-1.3cm}
\end{table}

\subsection{Layering-Specific Evaluation and User Study}
\label{sec:user_study}

Because standard metrics (e.g., SSIM, LPIPS) do not explicitly capture LVTON-specific challenges like occlusion and inner-layer preservation, we conducted a $30$-participant user study. Using a $5$-point Likert scale ($1$ = Strongly Disagree to $5$ = Strongly Agree), users evaluated our method against baselines across five criteria: 
(i) \textbf{\textit{Visual Realism}}: the generated image looks visually realistic and natural; 
(ii) \textbf{\textit{Instruction Following}}: the image correctly aligns with the given text instruction; 
(iii) \textbf{\textit{Occlusion Correctness}}: the occlusion relationships between garments are accurate (e.g., outer garments appropriately cover inner garments); 
(iv) \textbf{\textit{Inner-Layer Visibility}}: relevant inner-layer details (e.g., collars, sleeves, hems) are properly preserved; 
and (v) \textbf{\textit{Overall Layering Quality}}: the garment layering appears coherent and physically plausible. 

As shown in~\cref{table:user_study}, our approach consistently outperforms baselines. Most notably, it achieves significant improvements in the layering-specific metrics (Occlusion Correctness, Inner-Layer Visibility, and Overall Layering Quality), demonstrating superior capability in handling complex multi-layer interactions. To complement our human evaluation, we introduce a quantitative metric, Masked Preservation Score (MPS), that measures the structural and perceptual preservation of non-target regions. As detailed in Appendix E, our method improves upon the SOTA on MPS, increasing SSIM by $0.005$ and reducing LPIPS by $0.006$.


\begin{figure}[t]
    \centering
    \includegraphics[width=1.0\linewidth]{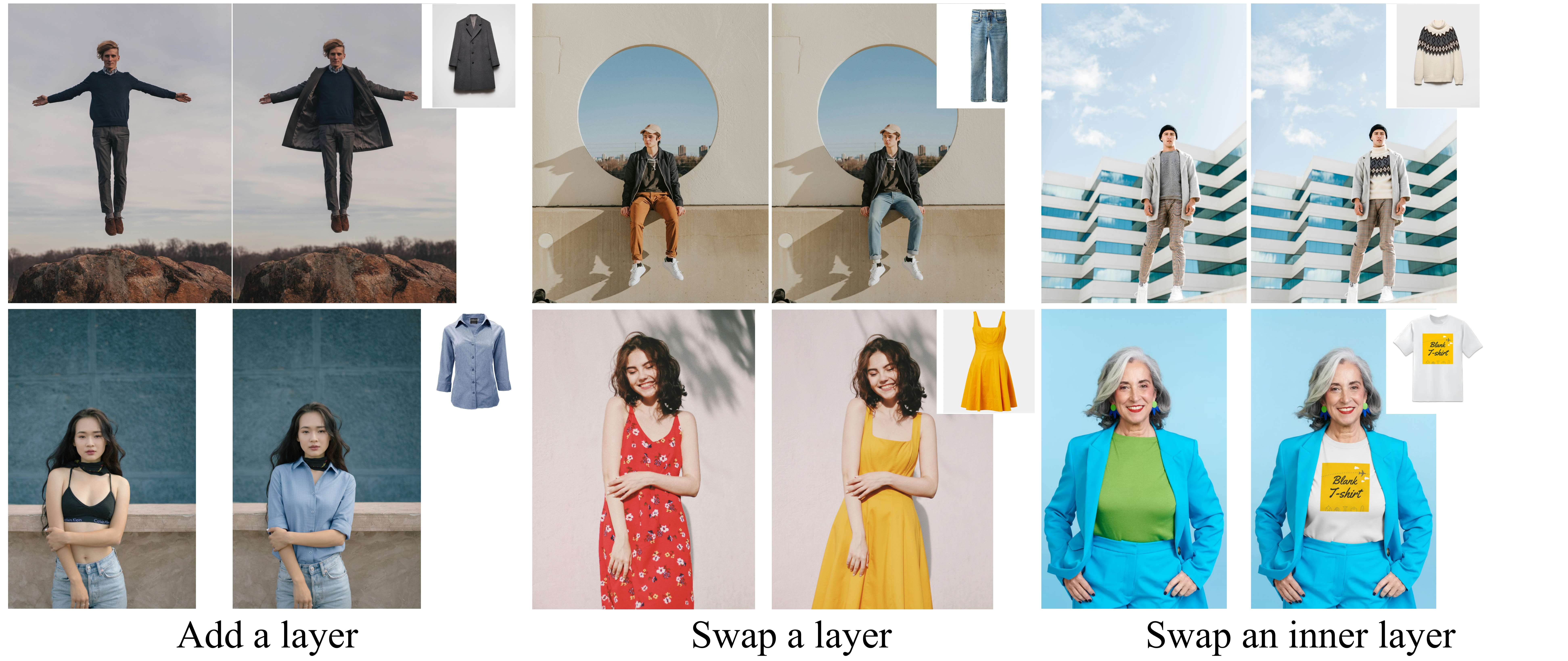}
    \vspace{-0.7cm}
    \caption{
        In-the-wild layering editing. Despite being fine-tuned on a strictly limited stage 2 dataset, our model successfully generalizes to complex, out-of-distribution scenarios sourced from the internet. It accurately executes specific compositional instructions (``add,'' ``swap,'' and inner-layer swaps) while naturally preserving the visual integrity of the non-target layers and overall high image fidelity.
    }
    \label{fig:wild_result}
    \vspace{-0.65cm}
\end{figure}

\vspace{-0.45cm}
\subsection{In-the-Wild Generalization}
\label{sec:in_the_wild}

Our approach demonstrates strong generalization to in-the-wild images sourced directly from the internet (\cref{fig:wild_result}). We evaluate the model on three types of layering edits --- adding an outer layer, swapping an existing garment, and replacing an inner layer --- across a wide range of unconstrained scenarios (e.g., outdoor scenes, varied poses, different genders, and diverse garment types). 

Despite the complexity of these in-the-wild images, the model maintains consistent subject identity and realistic shading. Furthermore, it accurately renders the target garments while flawlessly preserving the structural integrity and visibility of the non-target layers. These results confirm that our training paradigm is highly \textbf{data-efficient}: the model effectively learns specific layering logic from the limited stage 2 dataset and successfully generalizes to arbitrary, real-world dressing contexts without loss of visual fidelity.

While our method successfully handles various scenarios, it remains bounded by stage 2 data distribution. Improvements may require scaling the data collection pipeline to cover broader layering distribution. More details can be found in Appendix G.


\vspace{-0.5cm}
\begin{table}[h]
    \small
    \centering
    \caption{Ablation study on training components, evaluated on our layering dataset. This validates the necessity of two training stages and the temporal reversal augmentation. Ours achieves the best results, confirming our hypothesis. Best results are in \textbf{bold}.}
    \vspace{-0.3cm}
    \begin{tabular}{lcccc}
        \toprule
        Model & SSIM $\uparrow$ & LPIPS $\downarrow$ & FID $\downarrow$ & KID{\tiny $\times 10^{3}$} $\downarrow$ \\
        \midrule
        Stage 2-only       & $0.817$ & $0.192$ & $58.561$ & $13.242$ \\
        Stage 1-only       & $0.821$ & $0.183$ & $56.663$ & $12.286$ \\
        Ours  w/o Aug    & $0.832$ & $0.131$ & $49.353$ & $3.003$ \\
        \midrule
        Ours           & $\mathbf{0.843}$ & $\mathbf{0.127}$ & $\mathbf{48.957}$ & $\mathbf{1.463}$ \\
        \bottomrule
    \end{tabular}
    \label{table:ablation_table}
    \vspace{-0.8cm}
\end{table}

\subsection{Ablation Study}
\label{sec:ablation}

\paragraph{\textbf{Analysis of disentangled training stages.}} 
We evaluate three model variants on our LVTON dataset:
(1) \textbf{Stage 2-only}: trained from scratch only on our real-world layering data. This simulates ``direct training'' approach.
(2) \textbf{Stage 1-only}: our stage 1 model, which only learns ``swap'' logic. 
(3) \textbf{Ours w/o Aug}: our pipeline without the temporal reversal augmentation in stage 2.

As shown in~\cref{table:ablation_table}, the results strongly corroborate our hypothesis. First, Stage 2-only achieves the lowest quantitative scores, confirming that forcing the model to learn complex LVTON logic solely from scarce data is ineffective. Conversely, Stage 1-only exhibits a different failure mode. Since it is trained exclusively on ``swap'' (garment replacement) scenarios, it struggles to generalize to the ``add'' (layering) task. Finally, the performance drop in Ours w/o Aug validates the effectiveness of our temporal reversal augmentation.


\vspace{-0.3cm}
\paragraph{\textbf{Validation of general VTON priors.}} 
To confirm that stage 1 provides robust, generalizable priors, we evaluate Stage 1-only on VITON-HD and DressCode in a zero-shot setting. Crucially, our synthetic dataset was strictly deduplicated against these benchmarks to prevent data leakage and memorization. As~\cref{table:classical_result_table} shows, Stage 1-only achieves highly competitive results without seeing target domain data, even surpassing fully trained methods like IDM-VTON. This confirms that stage 1 builds a strong foundation for both general and layering VTON tasks.

\begin{figure}[t]
    \centering
    \includegraphics[width=0.55\linewidth]{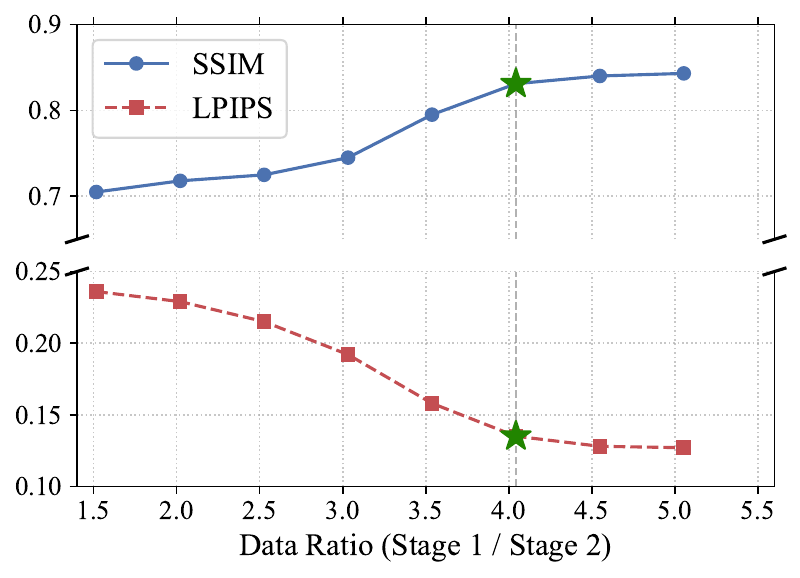}
    \vspace{-0.35cm}
    \caption{Ablation study on stage 1 data scale using the LVTON benchmark. The curves illustrate SSIM and LPIPS as a function of the data ratio between stage~1 and stage~2. Increasing the ratio yields consistent improvements, validating the efficacy of stage 1 priors. We adopt the saturation point of $4.0\times$ (green star) as the optimal configuration.}
    \label{fig:pretraining_ablation}
    \vspace{-0.7cm}
\end{figure}

\vspace{-0.3cm}
\paragraph{\textbf{Impact of stage 1 data scale.}} 
We investigate the relationship between the scale of the stage 1 synthetic data and the final LVTON performance. We train several models using synthetic data subsets of varying ratios relative to the fixed-size stage 2 data, and then fine-tune all models on the identical LVTON dataset. The results, plotted in~\cref{fig:pretraining_ablation}, reveal a clear and significant trend: performance (SSIM/LPIPS) improves substantially as the volume of stage 1 data increases. Notably, these performance gains begin to plateau around a $4.0\times$ data ratio (Stage~1~/~Stage~2), suggesting a point of saturation. This study strongly validates our hypothesis: the stage 1 training is crucial for building robust and generalizable VTON priors, enabling the model to achieve high fidelity when fine-tuning on the limited LVTON data.

\section{Limitations}
\label{sec:limitations}

\vspace{-0.1cm}
Our method has three limitations. 
(1) Stage 1 synthetic data quality is bound by upstream segmentation \cite{ravi2024sam} and inpainting \cite{batifol2025flux} models. Any artifacts or errors may be learned, degrading our VTON priors. 
(2) While the proposed method demonstrates strong generalization to in-the-wild layering scenarios, the limited scale and diversity of the stage 2 dataset restrict generalization to atypical or complex outfit interactions. 
(3) Our temporal reversal augmentation imperfectly simulates garment removal, missing real-world physical dynamics like inner-layer fabric pulling or hair disruption. Future data collection of authentic garment removal sequences can address this gap.

%
%
\bibliographystyle{splncs04}
\bibliography{main}

\newpage 
\appendix

\renewcommand{\thefigure}{A\arabic{figure}}
\renewcommand{\thetable}{A\arabic{table}}
\setcounter{figure}{0}
\setcounter{table}{0}

This supplementary material provides details on implementation specifics and deeper methodological analyses to support the main manuscript, concluding with an extensive gallery of qualitative results.

Specifically, the appendix is organized as:

\vspace{-0.2cm}
\paragraph{\textbf{Implementation and Dataset Details}}
\begin{itemize}
    \item \cref{sec:sup_implementation} outlines the implementation details, including training hyperparameters and optimization settings.
    
    \item \cref{sec:sup_dataset} elaborates on the automated pipelines used to construct the synthetic and real-world layering datasets. 
    
    \item \cref{sec:sup_dataset_analysis} follows with a statistical analysis of the distributions for both datasets. 
\end{itemize}

\vspace{-0.2cm}
\paragraph{\textbf{Methodological and Metric Analysis}}
\begin{itemize}
    \item \cref{sec:sup_qwen_comparison} analyzes the generative priors of our base model, Qwen-Image-Edit~\cite{wu2025qwen}. 
    
    \item \cref{sec:sup_metric} introduces the Masked Preservation Score (MPS) to explicitly quantify layering-specific preservation, detailing its diagnostic utility and scaling limitations.     
\end{itemize}

\vspace{-0.2cm}
\paragraph{\textbf{Discussions and Future Directions}}
\begin{itemize}
    \item \cref{sec:sup_3d_layering} contextualizes our 2D method within the broader landscape of 3D layered garment modeling.

    \item \cref{sec:sup_failure} discusses current limitations and representative failure cases, outlining directions for future research.
\end{itemize}

\vspace{-0.2cm}
\paragraph{\textbf{Additional Qualitative Results}}
\begin{itemize}
    \item \cref{sec:sup_layering} provides extensive qualitative results for the layering task on in-the-wild inputs. 
    
    \item \cref{sec:sup_traditional_vton} presents further comparisons on traditional VTON benchmarks.
\end{itemize}

\section{Implementation Details}
\label{sec:sup_implementation}

\paragraph{\textbf{Training paradigm and model setup.}} 
As introduced in the main paper, our method utilizes a two-stage training paradigm. We first train the model on our synthetic dataset (stage 1) to learn general VTON priors, and then fine-tune it on the real-world layering dataset (stage 2). We employ LoRA \cite{hu2022lora} to fine-tune the Qwen-Image-Edit model \cite{wu2025qwen}. For the LoRA adaptation, we set both the rank $r$ and the scaling factor $\alpha$ to $32$. We apply LoRA specifically to the key (K), query (Q), and value (V) projection modules within the diffusion transformer's attention layers. The core model architecture, training objective (standard denoising loss), and all optimization details remain identical across both stages. 

\vspace{-0.3cm}
\paragraph{\textbf{Optimization and hyperparameters.}} 
We employ the AdamW optimizer \cite{loshchilov2017decoupled} using its default hyperparameters (e.g., for $\beta_1, \beta_2, \epsilon$, and weight decay). We use a constant learning rate of $1 \times 10^{-4}$ throughout the entire training process, with no learning rate scheduler or warmup phase. All models are trained with a global batch size of $32$ using bf$16$ mixed precision. Further details on data preprocessing, including image resolution and augmentations, are described in~\cref{sec:sup_dataset}.

\vspace{-0.3cm}
\paragraph{\textbf{Training procedure and environment.}} 
The number of training steps varies slightly across benchmarks. For our LVTON benchmark, the VTON-prior-learning stage (stage 1) is trained for $20,000$ steps, followed by the fine-tuning stage (stage 2) for $5,000$ steps. For the traditional benchmarks \cite{choi2021viton, morelli2022dress}, stage 1 is also trained for $20,000$ steps, while stage 2 is fine-tuned for $3,000$ steps. We report results using the last checkpoint for all experiments and use a single NVIDIA H200 GPU. Our implementation is built using PyTorch (version \texttt{2.7.0+cu128}).
\section{Dataset Curation Details}
\label{sec:sup_dataset}

This section provides a detailed description of the dataset curation pipelines for the two-stage training paradigm introduced in the main paper. We detail the construction of the large-scale synthetic dataset for learning general VTON priors~(stage~1) and the specialized real-world dataset for learning layering knowledge~(stage~2).


\subsection{Stage 1: Synthetic Dataset}
\label{sec:sup_dataset_stage1}

The objective of this stage is to create a large-scale, mask-free, and pose-misaligned dataset to equip the model with robust ``deform-and-place'' priors.

\vspace{-0.3cm}
\paragraph{\textbf{Data sourcing and filtering.}} 
We source our raw video data from two primary streams: (1) publicly available fashion videos from platforms such as LTK, and (2) the TikTok dataset \cite{jafarian2021learning}. To ensure suitability for our task, all videos undergo a rigorous two-step filtering process.
\begin{itemize}
    \item \textbf{VLM pre-filtering:} we employ a Vision-Language Model (VLM), Qwen-2.5-VL (the $72$B variant) \cite{bai2025qwen2}, to automatically scan all videos. This VLM filters for two main criteria: (a) containing only a single person, and (b) assessing outfit stability. If the person's outfit remains constant throughout the video, the entire video is accepted. If the video contains multiple stable outfits, we do not discard it. Instead, we employ a VLM-based sliding-window detection technique (detailed in~\cref{sec:sup_dataset_stage2}) to partition the video into multiple distinct clips. Each resulting clip, containing one constant outfit, is then processed as an independent video source.
    \item \textbf{Manual review:} the VLM-filtered videos and the automatically generated clips are then manually reviewed by our team to guarantee $100\%$ adherence to these criteria, ensuring no multiple subjects are present.
\end{itemize}

\vspace{-0.3cm}
\paragraph{\textbf{Preliminary processing and frame selection.}} 
The filtered videos are first decomposed into individual frames at a rate of $5$ frames per second (fps). A significant challenge, particularly in the TikTok dataset \cite{jafarian2021learning}, is the presence of motion blur caused by rapid motion (e.g., dancing), which is detrimental to our VTON task. To mitigate this, we apply a blur detection algorithm based on the Laplacian operator. We compute the Laplacian variance for each frame and discard those with variance below a predefined threshold, ensuring our dataset consists of high-quality, sharp images.

\vspace{-0.3cm}
\paragraph{\textbf{Automated annotation and synthesis.}} 
After frame selection, we process the frames to generate the components needed for our synthetic pipeline.
\begin{itemize}
    \item \textbf{Base annotation:} we apply a pose estimator \cite{yang2023effective} and a segmentation model \cite{ravi2024sam} to extract the pose $P_i$, garment mask $M_i$, and segmented garment $G_i$ for every frame $F_i$.
    \item \textbf{Garment description:} the VLM \cite{bai2025qwen2} is used to generate a single, consistent text description $\mathcal{T}$ for the outfit in each video sequence.
    \item \textbf{Inpainting mask generation:} a key step in our process is creating an expanded mask for inpainting. Instead of segmenting only the garment, we also segment the person's arms or legs (depending on whether the garment is upper-body or lower-body) and merge them with $M_i$. This expanded mask provides the inpainting model \cite{batifol2025flux} with greater creative freedom, enabling more significant topological changes (e.g., transforming a T-shirt into a long-sleeve shirt or vice versa), thereby increasing data diversity. Since most layering actions occur on the upper body, we prioritize this region. When processing full-body frames, we apply a segmentation ratio of $0.7$ (upper body) to $0.3$ (lower body), cropping the frame or focusing the segmentation masks accordingly to capture the most relevant apparel items.
    \item \textbf{Synthesis prompt generation:} we utilize an LLM \cite{comanici2025gemini} to generate two sets of terms: (1) a set of ``garment descriptors'' (e.g., ``plaid,'' ``wool,'' ``dark blue'') and (2) a set of ``garment types'' (e.g., ``coat,'' ``cardigan,'' ``sweater''). A novel inpainting prompt is created by independently sampling one term from each set and combining them (e.g., ``a plaid coat'').
    \item \textbf{Image synthesis:} finally, for each frame $F_i$, we use the inpainting model \cite{batifol2025flux} guided by the expanded mask and a randomly generated synthesis prompt to create a new image $F_i^{\prime}$ featuring a novel outfit.
\end{itemize}

\begin{figure}[t]
    \centering
    \includegraphics[width=1.0\linewidth]{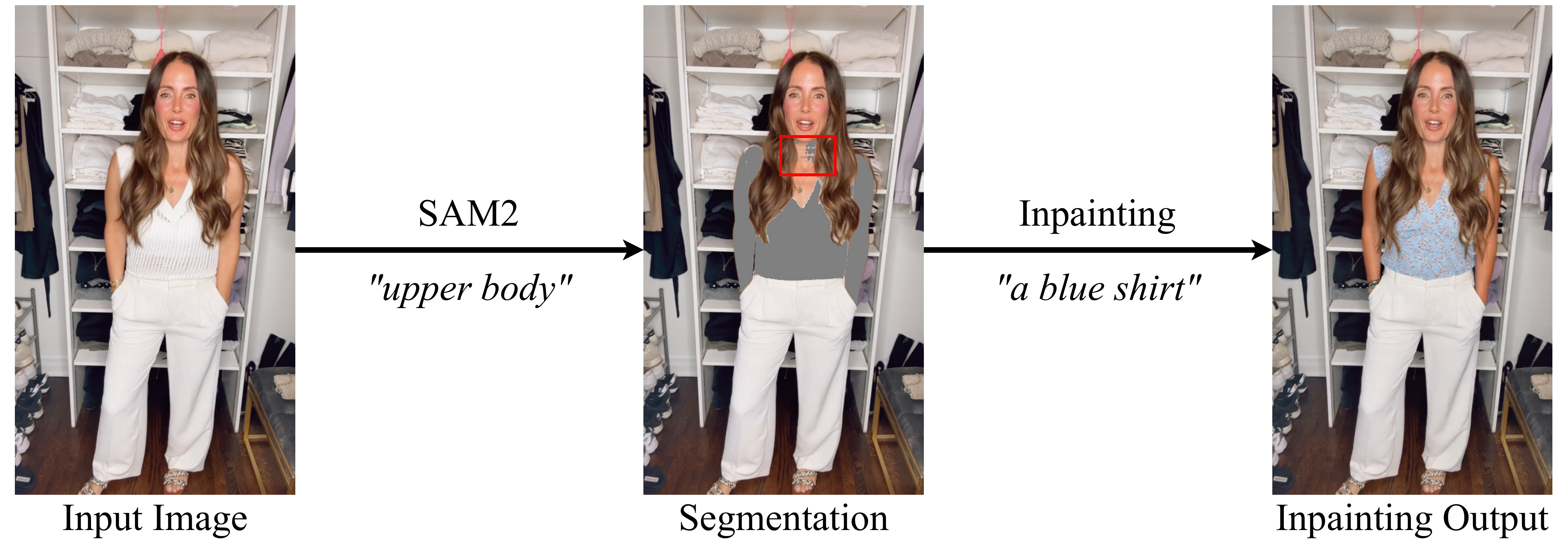}
    \vspace{-0.7cm}
    \caption{
        Illustration of stage 1 generation artifacts. The middle panel demonstrates an imprecise segmentation mask (red box) from SAM2~\cite{ravi2024sam} when given the ``upper body'' prompt. The right panel shows the final inpainting output, in which the model fails to strictly follow the ``a blue shirt'' prompt, producing unintended patterns rather than a solid color.
    }
    \label{fig:sup_stage1_failure}
    \vspace{-0.5cm}
\end{figure}

\vspace{-0.3cm}
\paragraph{\textbf{Training pair construction.}}
As detailed in the main paper, our training instances are constructed as $(\mathcal{I}_{M}, \mathcal{I}_{G}, \mathcal{I}_{P}, \mathcal{T}, \mathcal{I}_{T}) := (F_{i}^{\prime}, G_{j}, P_{j}, \mathcal{T}_{j}, F_{j})$, with the crucial constraint that $i \neq j$. To further enforce the ``deform-and-place'' challenge and prevent the model from learning trivial alignments, we introduce a ``dedicated garment frame'' strategy. For each video sequence, we designate one specific frame (e.g., $F_k$) to serve \textit{exclusively} as the garment source. The segmented garment $G_k$ and its description $\mathcal{T}_k$ from this frame are used as the garment input $\mathcal{I}_{G}$ and text prompt $\mathcal{T}$ for all training pairs generated from that video. The source person image $F_i^{\prime}$ and the target pose/image $(P_j, F_j)$ are then sampled from the remaining frames in the sequence (where $i, j \neq k$ and $i \neq j$). This strategy ensures that the input garment $G_k$ is almost always spatially misaligned with the target pose $P_j$, compelling the model to learn robust spatial deformation knowledge.

\vspace{-0.1cm}
\paragraph{\textbf{Stage 1 failure and mitigation strategy.}} 
While our automated data generation pipeline enables the efficient creation of a diverse dataset, we acknowledge the presence of inherent generation artifacts. As illustrated in~\cref{fig:sup_stage1_failure}, foundation models can occasionally fail: SAM2~\cite{ravi2024sam} may produce imprecise boundaries (e.g., an incorrect mask artifact near the neckline), and the subsequent inpainting model~\cite{batifol2025flux} may not strictly adhere to textual constraints (e.g., generating unintended patterns rather than a solid color).

Rather than employing heuristic mask refinement techniques (e.g., morphological dilation or erosion) or introducing complex auxiliary loss functions during training to combat texture degradation, we adopt a scale-driven mitigation strategy. Our approach relies on three key principles:

\vspace{-0.2cm}
\begin{itemize}
    \item \textbf{High Foundation Model Accuracy}: SAM2~\cite{ravi2024sam} and the chosen inpainting model~\cite{batifol2025flux} successfully process the vast majority of frames without significant artifacts.

    \item \textbf{Aggressive Discarding}: We prioritize dataset quality (precision) over yield (recall). Any generated pairs exhibiting noticeable segmentation failures or severe inpainting artifacts that degrade the final texture are strictly discarded.

    \item \textbf{Pipeline Scalability}: Because the data generation is fully automated and draws from in-the-wild videos, the pipeline is highly scalable. We can afford to aggressively filter out potentially flawed samples and easily generate replacements to meet our dataset size requirements.
\end{itemize}

We consider these occasional imperfections to be standard byproducts of utilizing foundation models. For future dataset scaling, implementing stricter, automated post-generation filtering protocols (e.g., using Vision-Language Models as evaluators) can further streamline the removal of these artifacts.

\vspace{-0.3cm}
\paragraph{\textbf{Addressing the synthetic-to-real domain gap.}} 
A natural concern when using automated inpainting is the potential domain gap between synthetic source garments and real-world target images. Despite this gap, training on these synthetic images is highly effective due to our disentangled two-stage paradigm. As established in the main manuscript, the primary objective of stage 1 is strictly to learn robust spatial deformation, alignment, and ``deform-and-place'' priors, not to master high-fidelity texture rendering. Any residual domain gap, such as minor inpainting artifacts or synthetic texture characteristics, is explicitly resolved in stage 2. Because stage 2 fine-tunes the model exclusively on high-quality, real-world layering data, it effectively acts as a domain adaptation step. The model retains the complex geometric priors learned from the synthetic data while calibrating its texture and shading generation to the real domain, a strategy empirically validated by our state-of-the-art results.


\subsection{Stage 2: Real-World Layering Dataset}
\label{sec:sup_dataset_stage2}

The objective of this stage is to curate a high-quality, real-world dataset to teach the model compositional reasoning for layering (e.g., ``add,'' ``remove,'' ``swap'').

\vspace{-0.3cm}
\paragraph{\textbf{Data sourcing and filtering.}} 
We download videos from YouTube using targeted keywords related to apparel layering, such as ``outfit for winter'' and ``how to layer for fall''. These videos undergo a strict manual filtering process. We only retain videos that meet three criteria: (1) feature a single person, (2) the person sequentially modifies their outfit, and (3) each transition involves a change of only one layer (e.g., adding a sweater, then adding a coat, rather than changing the entire outfit at once).

\vspace{-0.3cm}
\paragraph{\textbf{Outfit change detection and clustering.}} 
Manually identifying the exact frames where an outfit change occurs is prohibitively time-consuming. We observe that applying a VLM directly to a full video yields inaccurate temporal localization. Therefore, we develop a sliding window detection method. We define a window of size $W=6$ frames. Starting from the first frame, we feed all $W$ frames into the VLM \cite{bai2025qwen2} and prompt it to detect if an outfit change has occurred within this window. If no change is detected, the window slides forward. If a change is detected, we record the transition point and restart the sliding window process from the frame immediately following the detected change. This technique enables us to accurately partition the video's frame sequence into a series of temporal clusters $\{C_{1}, C_{2}, \cdots, C_{K}\}$, where each cluster $C_m$ contains a stable outfit configuration.

\vspace{-0.3cm}
\paragraph{\textbf{Automated annotation.}} 
Once the clusters are defined, we automate the annotation process.
\begin{itemize}
    \item \textbf{Pose and segmentation:} we first run the same pose estimator \cite{yang2023effective} and segmentation model \cite{ravi2024sam} on all frames to extract pose $P$ and initial segmentation masks $M$.
    \item \textbf{Transition description generation:} we sample one frame from an initial cluster $C_m$ and one from the subsequent cluster $C_{m+1}$ and feed both to the same VLM \cite{bai2025qwen2}. The VLM is prompted to generate a natural language description $\mathcal{T}_{m \rightarrow m + 1}$ describing the transition (e.g., ``add a gray sweater,'' ``swap the T-shirt for a blouse'').
    \item \textbf{VLM-guided segmentation:} the garment $G$ to be used as input is directly guided by the generated transition text $\mathcal{T}$. For example, if $\mathcal{T}$ is ``swap the sweater for a coat'', we segment the sweater ($G_i$) from frames in $C_m$ and the coat ($G_j$) from frames in $C_{m+1}$. If $\mathcal{T}$ is ``add a coat'', we only segment the coat ($G_j$) from frames in $C_{m+1}$. 
\end{itemize}

\vspace{-0.3cm}
\paragraph{\textbf{Forward pair construction.}} 
As described in the main paper, we construct the primary training tuples by sampling a source frame $F_i \in C_m$ and a target frame $F_j \in C_{m+1}$ from adjacent clusters. This forms the ``forward'' training instance: $(\mathcal{I}_{M}, \mathcal{I}_{G}, \mathcal{I}_{P}, \mathcal{T}, \mathcal{I}_{T}) := (F_{i}, G_{j}, P_{j}, \mathcal{T}_{m \rightarrow m + 1}, F_{j})$.

\vspace{-0.3cm}
\paragraph{\textbf{Temporal reversal augmentation.}} 
To address the scarcity of real-world layering data, we apply the temporal reversal augmentation technique mentioned in the main paper. For every forward instance representing an ``addition'' or ``swap'', we create a corresponding ``reverse'' instance: $(F_{j}, G_{i}, P_{i}, \mathcal{T}_{m+1 \rightarrow m}, F_{i})$. The reverse text prompt $\mathcal{T}_{m+1 \rightarrow m}$ is programmatically generated by applying lexical inversion rules to $\mathcal{T}_{m \rightarrow m + 1}$ (e.g., ``add \texttt{<garment>}'' becomes ``remove \texttt{<garment>}''). This augmentation strategy effectively doubles our fine-tuning dataset and provides explicit supervision for both garment addition and removal operations.



\section{Dataset Analysis}
\label{sec:sup_dataset_analysis}

\begin{figure}[t]
    \centering
    \includegraphics[width=1.0\linewidth]{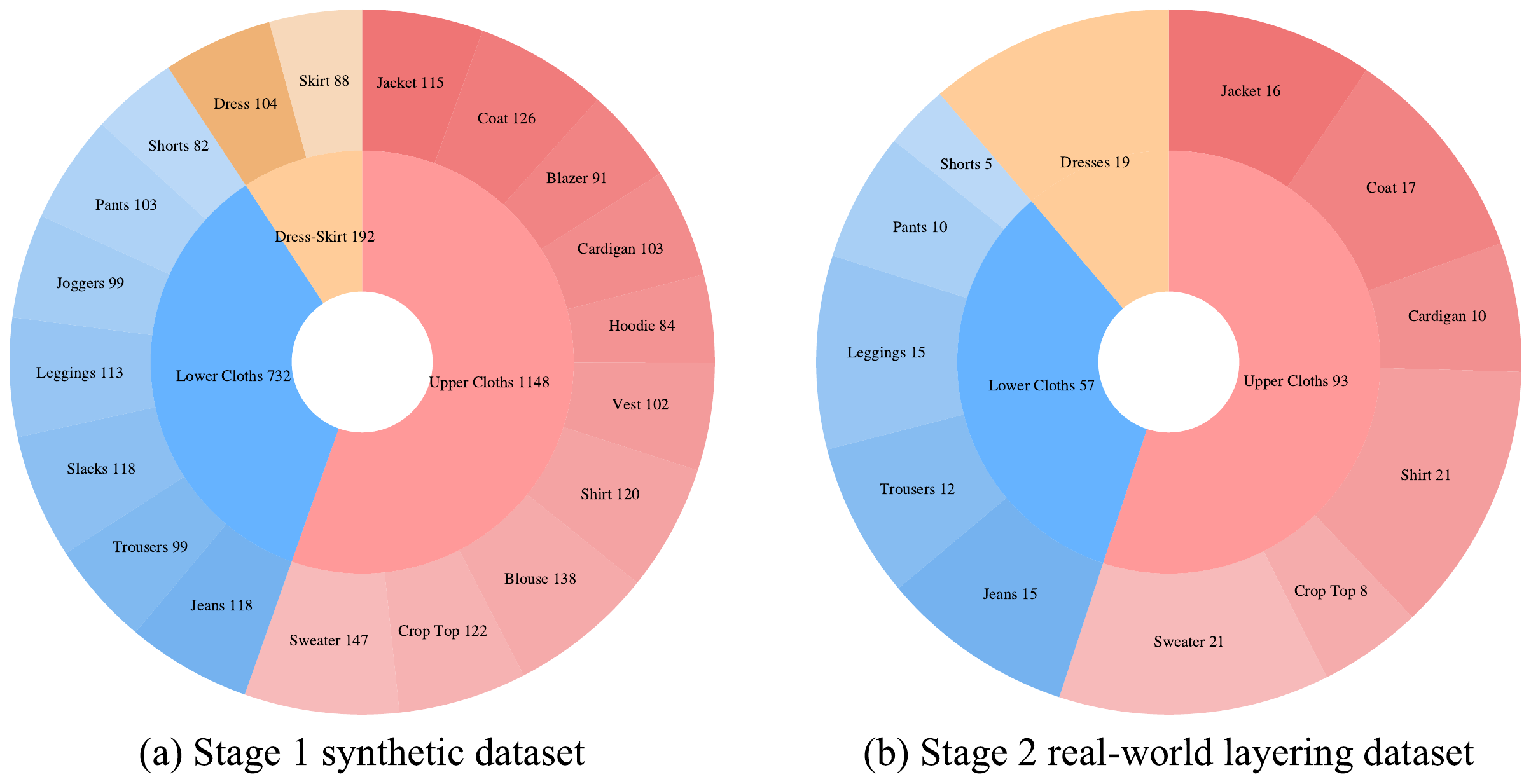}
    \vspace{-0.7cm}
    \caption{
        Distribution of garment categories in our two-stage datasets. (a) Stage 1 synthetic dataset. (b) Stage 2 real-world layering dataset. The inner ring shows broad categories (e.g., upper cloths, lower cloths), while the outer ring details fine-grained types. 
        \textbf{Clarification on counts:} the numbers in the outer ring represent the count of \textit{events}, not the final number of training pairs. We define an ``event'' as: (1) for stage 1, a unique video clip designated for inpainting with a specific garment prompt (e.g., ``jacket'' events). (2) for stage 2, a single, detected real-world outfit change transition (e.g., ``coat'' addition/swap events). 
        As detailed in~\cref{sec:sup_dataset_stage1} and~\cref{sec:sup_dataset_stage2}, we build training pairs within each event, which is why the total pair count is larger than event counts.
    }
    \label{fig:sup_dataset_analysis_figure}
    \vspace{-0.5cm}
\end{figure}

\vspace{-0.3cm}
\paragraph{\textbf{General statistics.}} 
Our training paradigm is supported by two distinct datasets. For stage 1 (general VTON priors), we constructed a large-scale synthetic dataset with $29,151$ training pairs derived from $362$ videos. For stage 2 (specific layering knowledge), we curated a specialized real-world dataset of $5,768$ training pairs and $532$ test pairs from $60$ videos, including the expansion resulting from our temporal reversal augmentation. Specifically, our stage 2 dataset contains only female subjects in general indoor scenes, reflecting a realistic e-commerce body-shape distribution without extreme outliers. The distribution of edit instructions is: $2,440$ ``add,'' $2,440$ ``remove,'' and $886$ ``swap.'' The lower proportion of ``swap'' is by design, since stage 1 already establishes ``swap'' priors. Garment category distributions are detailed in~\cref{fig:sup_dataset_analysis_figure}. All images in both datasets are resized to a resolution of~$896 \times 512$ pixels. 

\vspace{-0.3cm}
\paragraph{\textbf{Garment category analysis.}}
\cref{fig:sup_dataset_analysis_figure} visualizes the categorical distribution of both datasets. The counts shown in the figure represent ``events'' (i.e., synthetic clips for stage 1 or real-world transitions for stage 2) rather than the final counts of training pairs. This is because our pipeline builds training pairs within each unique event. 

The stage 1 dataset (\cref{fig:sup_dataset_analysis_figure}~(a)) is designed for breadth and diversity. It features a substantial, relatively balanced collection of garment events, including $1,148$ for upper clothing, $732$ for lower clothing, and $192$ for dress-skirt. The outer ring shows a wide range of fine-grained categories, including sweater ($147$), blouse ($138$), jeans ($118$), and pants ($103$). This large-scale dataset is crucial for teaching the model a robust and generalizable VTON prior before it encounters the more complex task of layering.

Conversely, the stage 2 dataset (\cref{fig:sup_dataset_analysis_figure}~(b)) has a smaller scale (e.g., $93$ upper cloth events and $57$ lower cloth events), reflecting the scarcity of high-quality, real-world layering data. The distribution is heavily skewed toward garments suitable for layering. This is evident in two ways: (1) the dataset is naturally dominated by upper cloths ($93$ vs. $57$), which is consistent with real-world layering behavior, and (2) the most prominent categories are all quintessential layering items, including shirt ($21$), sweater ($21$), jacket ($16$), and coat ($17$). This dataset is ideal for fine-tuning the model on the specific compositional knowledge required for LVTON.

\section{Analysis of Base Model Priors and Stage 1 Performance}
\label{sec:sup_qwen_comparison}


\vspace{-0.0cm}
\paragraph{\textbf{Base model capability.}} 
A natural question arises regarding the extent to which our model's performance on traditional VTON benchmarks stems from the robust priors memorized by the base model, Qwen-Image-Edit model~\cite{wu2025qwen}. To investigate this, we evaluate the zero-shot try-on capabilities of Qwen-Image-Edit against our Stage 1-only model and our complete two-stage model on the VITON-HD benchmark~\cite{choi2021viton}. The results are presented in~\cref{table:sup_qwen_comparison}. 

The base model possesses strong generative priors, achieving an FID of $6.901$ without any task-specific fine-tuning. However, it struggles with the precise spatial alignment and garment preservation required for virtual try-on, reflected in its lower SSIM ($0.869$) and higher LPIPS ($0.138$). 

\begin{table}[t]
    \centering
    \caption{
        Comparison of base model (i.e., Qwen-Image-Edit~\cite{wu2025qwen}) capabilities and our two-stage pipeline on VITON-HD~\cite{choi2021viton}. The base model shows strong generative priors but lacks the precise spatial alignment needed for VTON. Stage 1 training improves alignment (Stage 1-only's improvements in SSIM, LPIPS, and FID) but introduces a minor synthetic-to-real domain gap penalized by KID. Stage 2 effectively resolves this domain gap via domain adaptation, achieving the best performance across all metrics.
    }
    \vspace{-0.2cm}
    \begin{tabular}{lccccc}
        \toprule
        Method & SSIM $\uparrow$ & LPIPS $\downarrow$ & FID $\downarrow$ & KID{\tiny $\times 10^{3}$} $\downarrow$ \\
        
        \midrule

        Previous SOTA & $0.899$ & $0.057$ & $4.731$ & $0.190$ \\
        
        Qwen-Image-Edit~\cite{wu2025qwen} & $0.869$ & $0.138$ & $6.901$ & $1.180$ \\

        \midrule

        Stage 1-only & $0.887$ & $0.086$ & $5.235$ & $1.550$ \\
        
        Ours & $\textbf{0.928}$ & $\textbf{0.055}$ & $\textbf{3.235}$ & $\textbf{0.200}$  \\
        
        \bottomrule
    \end{tabular}
    \label{table:sup_qwen_comparison}
    \vspace{-0.45cm}
\end{table}


\vspace{-0.0cm}
\paragraph{\textbf{The impact of stage 1.}} 
When evaluating our Stage 1-only model, we observe significant improvements in SSIM ($0.887$), LPIPS ($0.086$), and FID ($5.235$) compared with the base model. This confirms that our stage 1 synthetic dataset successfully forces the model to learn the ``deform-and-place'' spatial priors that the base model inherently lacks.

\vspace{-0.0cm}
\paragraph{\textbf{Metric discrepancies.}} 
Notably, while the Stage 1-only model improves across most metrics, its KID score ($1.550$) is slightly worse than that of the base model ($1.180$). We argue that this discrepancy is a direct result of the domain gap between our automated synthetic training data and the high-resolution, real-world studio images of the VITON-HD benchmark. Because stage 1 pairs are synthesized by upstream models, they occasionally exhibit minor inpainting artifacts (see~``Stage~1~failure and mitigation strategy'' in~\cref{sec:sup_dataset_stage1}). While FID (which relies on Gaussian assumptions) rewards overall improvements in structural alignment, KID is a non-parametric metric that is highly sensitive to fine-grained, synthetic distributional shifts. The base model, pre-trained on massive real-world data, exhibits a slightly more ``natural'' and unbiased feature distribution, according to KID.


\vspace{-0.0cm}
\paragraph{\textbf{Resolution via stage 2.}} 
This metric anomaly further validates our disentangled two-stage paradigm. While stage 1 is crucial for establishing spatial deformation priors, it introduces a minor synthetic-to-real domain gap. Our stage 2 fine-tuning acts as an effective domain adaptation step (see ``Addressing the synthetic-to-real domain gap'' in~\cref{sec:sup_dataset_stage1}). As shown in~\cref{table:sup_qwen_comparison}, applying stage 2 resolves this gap, reducing the KID to $0.200$. 
\begin{table}[t]
    \centering
    \caption{
        Masked Preservation Score (MPS) on layering sequences. Evaluated specifically on ``add a garment'' tasks, our method achieves the best localized preservation of non-target inner layers and extremities.
    }
    \vspace{-0.3cm}
    \begin{tabular}{l c@{\hspace{0.1cm}}c c c@{\hspace{0.1cm}}c c}
        \toprule
         Model & && MPS-SSIM $\uparrow$ & && MPS-LPIPS $\downarrow$ \\
        
        \midrule
        
        OmniTry \cite{feng2025omnitry}          & && $0.611$ & && $0.425$ \\
        
        Any2AnyTryon \cite{guo2025any2anytryon} & && $0.579$ & && $0.418$ \\
        
        Nano Banana \cite{comanici2025gemini}   & && $0.744$ & && $0.295$ \\

        HunyuanImage-3.0-Instruct \cite{cao2025hunyuanimage}   & && $0.750$ & && $0.292$ \\
        
        Ours                                    & && $\mathbf{0.755}$ & && $\mathbf{0.286}$ \\
        
        \bottomrule
    \end{tabular}
    \label{table:sup_mask_metric}
    \vspace{-0.3cm}
\end{table}

\begin{figure}[h]
    \centering
    \includegraphics[width=1.0\linewidth]{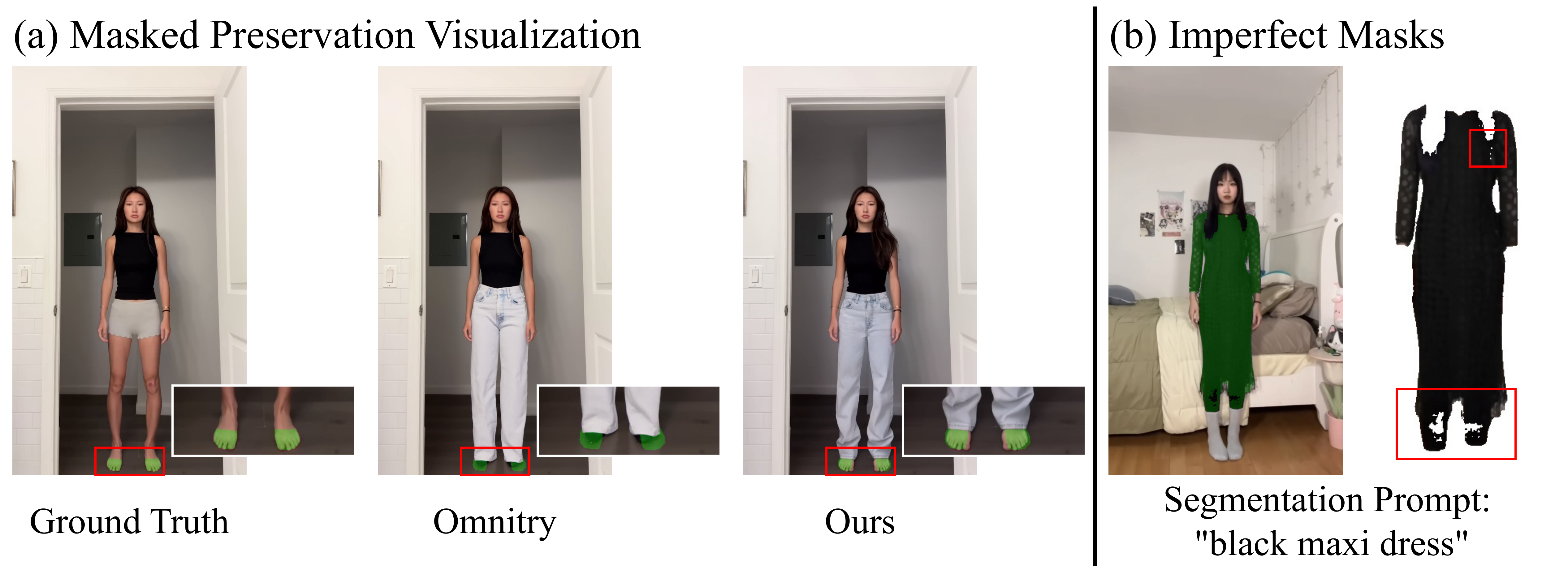}
    \vspace{-0.7cm}
    \caption{
        Exploring the Masked Preservation Score (MPS). \textbf{(a) Masked Preservation Visualization:} By isolating unchanged regions (e.g., the feet), we can compute localized metrics to reveal spatial displacements that global metrics miss. Our method preserves structural alignment much better than OmniTry. \textbf{(b) Imperfect Masks:} A key limitation of scaling MPS is its reliance on segmentation models, which can produce inaccurate evaluation boundaries (e.g., failing to correctly isolate the target region for a ``black maxi dress'').
    }
    \label{fig:sup_metric}
    \vspace{-0.5cm}
\end{figure}

\section{Exploring a Masked Preservation Score for Layering}
\label{sec:sup_metric}

\paragraph{\textbf{Motivation.}} 
While standard global metrics (e.g., SSIM~\cite{wang2004image}, LPIPS~\cite{zhang2018unreasonable}) provide a holistic view of generation quality, they inherently dilute the measurement of layering-specific performance by averaging over massive background and target-garment regions. As highlighted by our human evaluation (main paper Sec.~4.3), LVTON requires the strict preservation of non-target regions (e.g.,~inner~shirts, lower-body extremities). To provide an automated, quantitative counterpart to our human study's ``Inner-Layer Visibility'' criterion, we propose the \textbf{Masked~Preservation~Score~(MPS)}.

\vspace{-0.2cm}
\paragraph{\textbf{Calculation.}} 
MPS isolates unchanged regions using segmentation~\cite{ravi2024sam}. Given an original image and a generated image, we first extract the mask of the new target garment ($M_{\text{target}}$) by using the item description from the edit instruction (e.g., ``add a black maxi dress''). Then we extract the mask of the person's entire body ($M_{\text{body}}$). The evaluation region is defined as the strictly unchanged pixels: $M_{\text{unchanged}} = M_{\text{body}} \setminus M_{\text{target}}$. 

We then compute traditional metrics (i.e., SSIM~\cite{wang2004image}, LPIPS~\cite{zhang2018unreasonable}) exclusively within $M_{\text{unchanged}}$. For MPS-SSIM, we compute the global SSIM map and average only the values within the mask. For MPS-LPIPS, we isolate the unchanged regions by zeroing out the rest of the image before passing it to the VGG network~\cite{simonyan2014very}. Crucially, because standard LPIPS averages perceptual distance over the entire spatial dimension ($H \times W$), the raw score is heavily diluted by the zeroed-out areas. To correct this, we apply an area-penalty normalization, scaling the output by $\frac{H \times W}{|M_{\text{unchanged}}|}$, ensuring the metric accurately reflects the perceptual distance of the valid pixels alone.

\vspace{-0.2cm}
\paragraph{\textbf{Results and analysis.}} 
We evaluated MPS at scale across all baseline methods using our ``add a garment'' test subset, as this operation inherently requires preserving the original base outfit. As reported in~\cref{table:sup_mask_metric}, our method achieves the best MPS-SSIM and MPS-LPIPS, quantitatively validating its superior ability to preserve non-target details. This statistical advantage can be visually identified in~\cref{fig:sup_metric}~(a), where our method maintains structural alignment of the subject's extremities, whereas the baseline, OmniTry~\cite{feng2025omnitry}, suffers from severe spatial displacement. However, our method still introduces a slight displacement in the exact foot positioning compared to the original image, highlighting an area for future refinement in high-frequency identity preservation.

\vspace{-0.2cm}
\paragraph{\textbf{Limitations of MPS.}} 
While MPS is an effective diagnostic tool, its precision is upper-bounded by the accuracy of the segmentation model. In practice, segmentation models occasionally fail to capture precise boundaries or misinterpret complex layer interactions, leading to flawed evaluation masks (e.g., in~\cref{fig:sup_metric}~(b), the segmentation for a ``black maxi dress'' is incomplete). Simple morphological operations (e.g., dilation or erosion) cannot uniformly correct these errors without degrading initially accurate masks. Consequently, systematic noise is present in MPS calculations. Therefore, we present MPS as a strong supplementary diagnostic metric while relying on the human user study as the primary gold standard for overall layering evaluation.

\section{3D Layered Garment Modeling}
\label{sec:sup_3d_layering}

While our proposed LVTON method operates in the 2D image domain, a parallel and highly relevant line of research explores outfit layering through 3D representations. 

Recent advancements have made significant strides in reconstructing and simulating layered apparel. For instance, \cite{aggarwal2022layered} tackles the inverse problem of extracting distinct 3D implicit layers from a flat 2D image. In physics-based simulation, \cite{lee2023clothcombo} explicitly models the complex inter-cloth interactions, friction, and collision dynamics required for the realistic draping of multi-layered 3D garment meshes. More recently, \cite{xu2026layergs} has introduced techniques for decomposing layered 3D human avatars using 2D Gaussian Splatting~\cite{huang20242d}, enabling the inpainting and editing of individual layers within a spatially consistent 3D context.

Although our current 2D pipeline demonstrates strong empirical priors for spatial deformation, shading, and occlusion, it inherently lacks explicit geometric reasoning. This presents several exciting avenues for future work:

\vspace{-0.2cm}
\begin{itemize}
    \item \textbf{Physics-Guided Diffusion}: Incorporating physics-based simulation as additional conditioning for the diffusion model could resolve highly complex topological interactions. This would help generative models accurately render volumetric constraints, such as the bulging of a thick sweater beneath a tight leather jacket.

    \item \textbf{3D-Aware Layering}: Bridging our 2D generative flexibility with 3D physical accuracy could enable multi-view consistent LVTON. By lifting our 2D generated outfit layers into 3D implicit representations or Gaussian Splats, future systems could allow users to seamlessly rotate and view sequentially layered garments from any angle.
\end{itemize}
\vspace{-0.7cm}

\begin{figure}[h]
    \centering
    \includegraphics[width=1.0\linewidth]{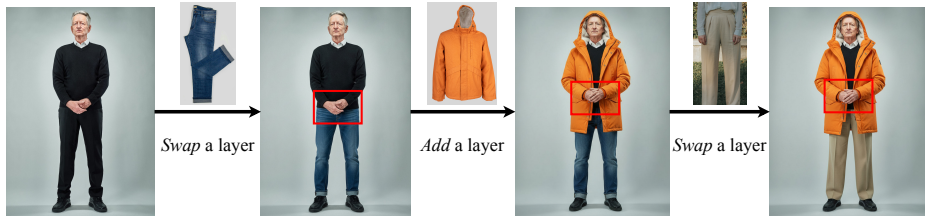}
    \vspace{-0.7cm}
    \caption{
        Failure case in hand pose preservation. While our method successfully performs a sequence of try-on operations (swap a layer, add a layer, swap a layer), it fails when adding the orange winter coat. The model unfaithfully alters the man's hand pose and hand positions, as highlighted by the red bounding boxes. This suggests a limitation in maintaining subtle, high-frequency identity details during complex layer manipulation.
    }
    \label{fig:sup_failure_figure}
    \vspace{-0.9cm}
\end{figure}

\section{Failure Case and Potential Directions for Improvements}
\label{sec:sup_failure}

Despite its overall success in complex layering sequences, our method exhibits certain limitations stemming from data scale and the inherent constraints of 2D diffusion models.

\vspace{-0.3cm}
\paragraph{\textbf{Data distribution and demographic bounds.}} 
While the stage 2 dataset efficiently teaches composition logic, its limited scale of $5,768$ training pairs cannot cover all real-world layering permutations. Specifically, the current dataset predominantly features female subjects in general indoor scenes. Consequently, the model may struggle to perfectly generalize to out-of-distribution scenarios, such as diverse male body types, highly atypical layering combinations, or extreme outdoor lighting conditions. Future improvements will require scaling our automated video-mining pipeline to actively collect and curate a vastly more diverse set of real-world layering interactions.

\vspace{-0.3cm}
\paragraph{\textbf{Volumetric and geometric constraints.}} 
Our approach lacks explicit geometric reasoning because it operates entirely in the 2D image domain. This limitation becomes apparent during complex physical interactions, such as attempting to layer a tight, restrictive outer garment over a bulky inner layer (e.g.,~a~thick winter sweater). In such cases, the 2D model cannot render the physical bulging or volume constraints. Future research could explore simulation-based methods as an additional conditioning. More details can be found in~\cref{sec:sup_3d_layering}. 

\vspace{-0.3cm}
\paragraph{\textbf{High-frequency identity preservation.}} 
Finally, the model occasionally struggles to consistently preserve subtle, high-frequency identity characteristics during complex manipulation. \cref{fig:sup_failure_figure} illustrates a representative failure case where the model struggles to consistently preserve the original identity and pose. During the ``add a layer'' operation, where the orange winter coat is applied to the model (moving from the second to the third image), the system incorrectly and noticeably alters the man's hand pose and placement. The red bounding boxes highlight this involuntary change. This suggests that while the model handles broad garment manipulation effectively, its capacity for maintaining subtle, high-frequency identity characteristics, such as the exact configuration and position of hands, is sometimes compromised when introducing new, complex garment layers. Future work should focus on exploring advanced mechanisms for robust identity preservation that are invariant to pose and layer manipulation, thereby ensuring a more faithful generation of the original human subject, especially in expressive areas such as the hands and face.

\section{Additional Layering Results}
\label{sec:sup_layering}

In this section, we provide additional qualitative results for the layering virtual try-on task, supplementing the examples presented in the main manuscript. As shown in~\cref{fig:sup_layering_result_figure}, these comparisons use a more challenging condition: the garment inputs are derived from in-the-wild segmentations rather than from clean, catalog-style product images. This setting tests the model's robustness to variations in lighting, wrinkles, and non-canonical poses inherent in segmented garments.

We observe that this scenario poses a significant challenge for existing baseline methods. Notably, Nano Banana~\cite{comanici2025gemini} consistently fails to perform try-on, often defaulting to rendering the original person's image without the new garment. We attribute this severe failure to a domain gap: the model was likely trained exclusively on clean product images and thus fails to generalize to in-the-wild segmentation domains. OmniTry~\cite{feng2025omnitry} and Any2AnyTryon~\cite{guo2025any2anytryon} also struggle, producing significant visual artifacts as highlighted in the red boxes. In contrast, our method demonstrates superior robustness, successfully handling these challenging segmented inputs to generate plausible and high-fidelity layered try-on results.
\section{Additional Traditional VTON Results}
\label{sec:sup_traditional_vton}

In this section, we present further visual results to supplement the traditional VTON evaluation discussed in the main paper. \cref{fig:sup_traditional_vton_results_1} and~\cref{fig:sup_traditional_vton_results_2} showcase additional qualitative comparisons on both the VITON-HD~\cite{choi2021viton} and DressCode~\cite{morelli2022dress} datasets. These results align with our quantitative findings, confirming that our proposed two-stage model consistently outperforms baseline methods. Specifically, our method maintains high structural coherence and preserves garment details, even in challenging scenarios where baseline methods tend to produce noticeable artifacts.


\afterpage{
    \clearpage
    \begin{figure}[p]
        \centering
        \includegraphics[width=1.0\linewidth]{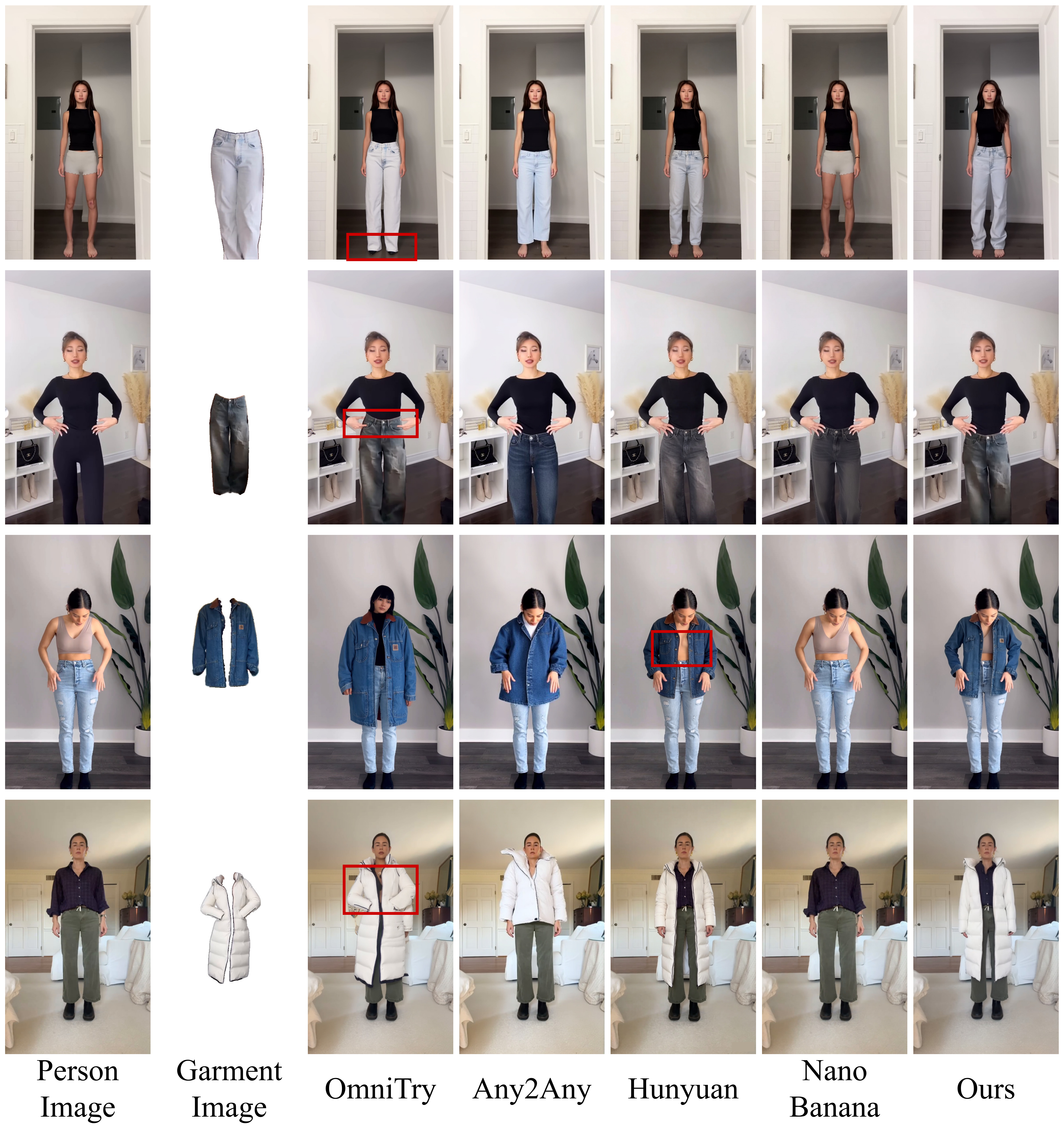}
        \caption{
            Additional qualitative comparisons for layering virtual try-on using garment segmentations. Different from the results in the main paper, we use challenging garment inputs derived from segmentations of in-the-wild images. Note that Nano Banana~\cite{comanici2025gemini} often fails on these inputs, which we attribute to a domain gap. OmniTry~\cite{feng2025omnitry}, Any2AnyTryon~\cite{guo2025any2anytryon}, and HunyuanImage-3.0-Instruct~\cite{cao2025hunyuanimage} also produce severe artifacts (highlighted by red boxes). Our method demonstrates significantly greater robustness to this data domain, generating plausible layered results.
        }
        \label{fig:sup_layering_result_figure}
    \end{figure}
    \clearpage
}


\afterpage{
    \clearpage
    \begin{figure}[p]
        \centering
        \includegraphics[width=1.0\linewidth]{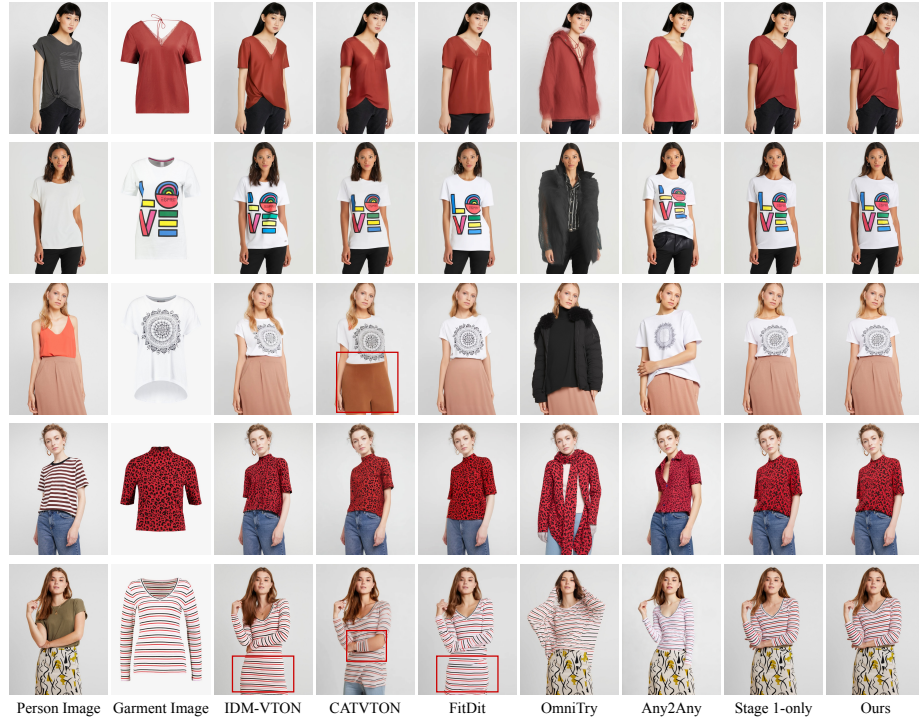}
        \caption{
            Additional qualitative comparisons on VITON-HD~\cite{choi2021viton} and DressCode~\cite{morelli2022dress} datasets. Similar to the observations in the main paper, our method generates high-fidelity try-on results that preserve garment details and body structure. Red boxes indicate artifacts present in the baseline results, which our method effectively addresses.
        }
        \label{fig:sup_traditional_vton_results_1}
    \end{figure}
    \clearpage
}


\afterpage{
    \clearpage
    \begin{figure}[p]
        \centering
        \includegraphics[width=1.0\linewidth]{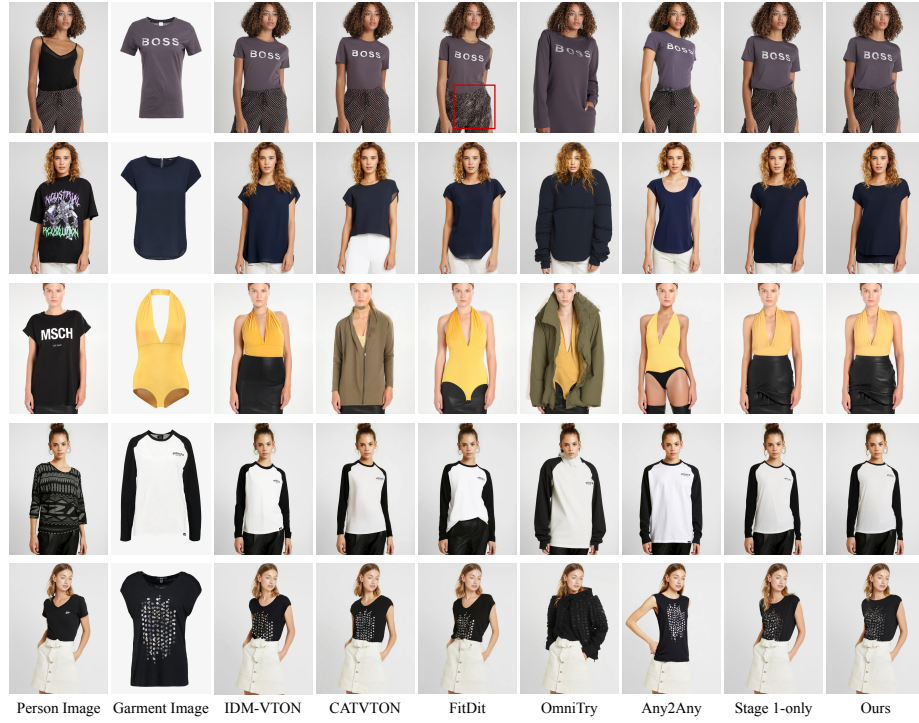}
        \caption{
            More qualitative comparisons. Further examples from the benchmark datasets demonstrate the generalization ability of our method. Consistent with~\cref{fig:sup_traditional_vton_results_1}, our approach maintains high fidelity across various garment types and poses.
        }
        \label{fig:sup_traditional_vton_results_2}
    \end{figure}
    \clearpage
}


\end{document}